\pdfoutput=1
\documentclass[11pt]{article}

\usepackage[]{acl}

\usepackage{times}
\usepackage{latexsym}
\usepackage{booktabs}
\usepackage{enumitem}
\usepackage{graphicx}
\usepackage{subcaption}
\usepackage[normalem]{ulem}
\usepackage{longtable}
\usepackage[most]{tcolorbox}
\usepackage{listings} % This package is used to format code
\usepackage{xcolor} % Required for setting colors
\usepackage{pdfpages}
\usepackage{lipsum}

\useunder{\uline}{\ul}{}

\let\oldquote\quote
\let\endoldquote\endquote
\renewenvironment{quote}[2][]
  {\if\relax\detokenize{#1}\relax
     \def\quoteauthor{#2}%
   \else
     \def\quoteauthor{#2~---~#1}%
   \fi
   \oldquote}
  {\par\nobreak\smallskip\hfill(\quoteauthor)%
   \endoldquote\addvspace{\bigskipamount}}

\newcommand\blfootnote[1]{%
  \begingroup
  \renewcommand\thefootnote{}\footnote{#1}%
  \addtocounter{footnote}{-1}%
  \endgroup
}

\usepackage[T1]{fontenc}
\usepackage[utf8]{inputenc}

\usepackage{microtype}

\usepackage{inconsolata}

\usepackage{multirow}

\usepackage{stfloats}

\definecolor{codegreen}{rgb}{0,0.6,0}
\definecolor{codegray}{rgb}{0.5,0.5,0.5}
\definecolor{codepurple}{rgb}{0.58,0,0.82}
\definecolor{backcolour}{rgb}{0.95,0.95,0.92}

\lstdefinestyle{mystyle}{
    backgroundcolor=\color{backcolour},   
    commentstyle=\color{codegreen},
    keywordstyle=\color{magenta},
    numberstyle=\tiny\color{codegray},
    stringstyle=\color{codepurple},
    basicstyle=\footnotesize\ttfamily,
    breakatwhitespace=true,         
    breaklines=true,                 
    captionpos=b,                    
    keepspaces=true,                 
    numbers=none, % No line numbers
    numbersep=5pt,                  
    showspaces=false,                
    showstringspaces=false,
    showtabs=false,                  
    tabsize=2,
    xleftmargin=0pt,
    xrightmargin=18pt,
    frame=none, % No frame around the code
    framexleftmargin=0pt,
    framexrightmargin=18pt,
    framextopmargin=0pt,
    framexbottommargin=0pt,
    showstringspaces=false,
    upquote=true,
}

\title{Measuring Psychological Depth in Language Models}

 \author{Fabrice Harel-Canada \quad  Hanyu Zhou \quad Sreya Muppalla \quad Zeynep Yildiz \\ \quad \textbf{Miryung Kim} \quad \textbf{Amit Sahai}\textsuperscript{‡} \quad \textbf{Nanyun Peng}\textsuperscript{‡} \\[7pt]
         University of California, Los Angeles\\[3pt]
         {
         \texttt{fabricehc@cs.ucla.edu}
         }
         }

\begin{document}
\maketitle

\begin{abstract}
Evaluations of creative stories generated by large language models (LLMs) often focus on objective properties of the text, such as its style, coherence, and diversity. While these metrics are indispensable, they do not speak to a story's subjective, psychological impact from a reader's perspective. We introduce the Psychological Depth Scale (PDS), a novel framework rooted in literary theory that measures an LLM's ability to produce authentic and narratively complex stories that provoke emotion, empathy, and engagement. We empirically validate our framework by showing that humans can consistently evaluate stories based on PDS ($0.72$ Krippendorff's alpha). We also explore techniques for automating the PDS to easily scale future analyses. GPT-4o, combined with a novel Mixture-of-Personas (MoP) prompting strategy, achieves an average Spearman correlation of $0.51$ with human judgment while Llama-3-70B with constrained decoding scores as high as $0.68$ for empathy. Finally, we compared the depth of stories authored by both humans and LLMs. Surprisingly, GPT-4 stories either surpassed or were statistically indistinguishable from highly-rated human-written stories sourced from Reddit. By shifting the focus from text to reader, the Psychological Depth Scale is a validated, automated, and systematic means of measuring the capacity of LLMs to connect with humans through the stories they tell. 
\end{abstract}

\blfootnote{\textsuperscript{‡} Equal advisory role. \\ Our code and data is available at \url{https://github.com/PlusLabNLP/psychdepth}.}
\section{Introduction}

Stories play a crucial role in our understanding of ourselves and the world around us  \cite{story_importance1, story_importance2}. As large language models (LLMs) are increasingly deployed in narrative design and creation, their growing impact on how stories are told calls for a deeper understanding of their narrative power. Current evaluations of LLM-generated stories often focus on objective properties of the text such as discourse structure \cite{liu-etal-2024-unlocking}, fluency \cite{creative-eval}, style \cite{weaver},  creativity \cite{art_or_artifice}, diversity \cite{gem}, toxicity and bias \cite{wang2023decodingtrust}. However, it is crucial to extend these evaluations to accommodate the subjective, psychological impact stories have on readers. While some studies have recently explored aspects like empathy \cite{empathy2, empathy3} and engagement \cite{engagement, endex}, they do not fully capture the multifaceted and interconnected nature of the reading experience.

% as Figure \ref{fig:teaser} illustrates

\begin{figure}
    \centering
    \includegraphics[width=\columnwidth]{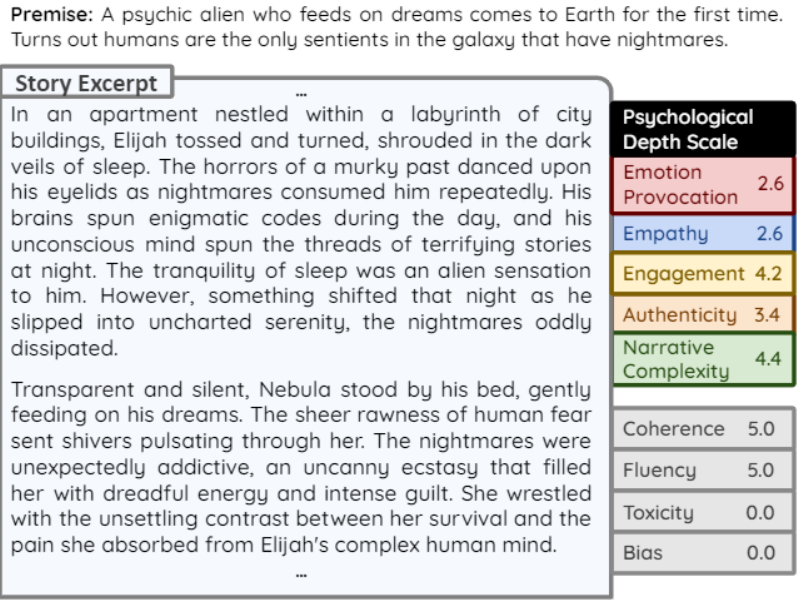}
    \caption{In this GPT-4 story, the psychological depth scale highlights strengths and weaknesses contributing to the overall reader experience, providing additional quality signals over traditional metrics more likely to saturate. Scores are normalized 1-5 for comparison.}
    \label{fig:teaser}
    \vspace{-5mm}
\end{figure}

Recognizing this gap, our study introduces a novel approach to measuring the \emph{psychological depth} of short stories. We present the Psychological Depth Scale (PDS), drawing inspiration from two literary theory frameworks: reader-response criticism \cite{reader_response} and text world theory \cite{text_world}. Reader-response criticism emphasizes the reader's subjective experience, while text world theory examines how readers cognitively construct a nuanced and realistic mental model of a story. Using related search terms, we conducted an extensive literature review of 95 peer-reviewed articles and books, identified 143 different evaluation criteria, and merged many of the broader themes into five key metacomponents: empathy, engagement, emotion provocation, authenticity, and narrative complexity. By recognizing the dual roles of authors and readers in creating and interpreting narratives, we aim to offer a more streamlined and comprehensive framework for assessing the psychological depth of creative content (Figure \ref{fig:teaser}).

\begin{figure*}
    \centering
    \vspace{-32mm}
    \includegraphics[width=\textwidth]{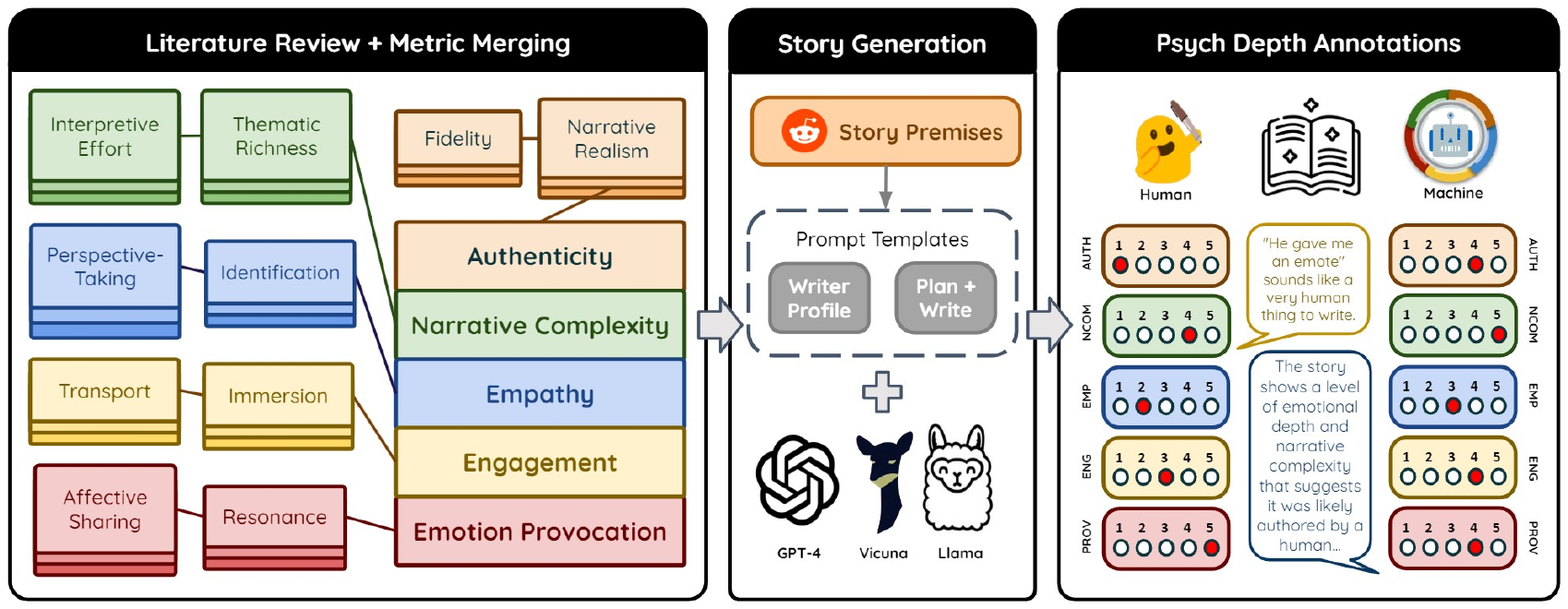}
    \vspace{-32mm}
    \caption{Overview of our approach to developing and validating the Psychological Depth Scale. We merge related metrics from an extensive survey of literary theory and reader-response analysis, then generate deep stories using LLMs, and finally compare annotations from both human evaluators and automated systems across five key dimensions: authenticity, narrative complexity, empathy, engagement, and emotion provocation.}
    \label{fig:psychdepth_overview}
    \vspace{-3mm}
\end{figure*}

To empirically validate the PDS, we conducted a study involving non-expert humans, who are increasingly engaging with creative content generated by LLMs. We enlisted five undergraduate students from UCLA to analyze a dataset of 97 stories authored by humans and five contemporary LLMs. The raters provided psychological depth ratings and predictions on human or machine authorship, with detailed justifications. Our approach, illustrated in Figure \ref{fig:psychdepth_overview}, addresses three critical research questions to explore the intersection of LLMs and the psychological nuances of creative writing.

\begin{itemize}[align=left, labelwidth=0.5cm, labelsep=0.15cm, leftmargin=0.75cm, noitemsep, nolistsep]
    \item[\textbf{RQ1.}] \emph{How consistently can well-informed humans judge psychological depth?} The Psychological Depth Scale achieved an average Krippendorff's alpha of $0.72$, indicating a significant level of agreement among raters and affirming its validity as a reliable instrument for assessing fictional short stories.
    \item[\textbf{RQ2.}] \emph{To what extent can psychological depth be measured automatically?} Leveraging our novel Mixture-of-Personas prompting strategy, GPT-4o achieved a Spearman correlation of $0.51$ with human judgment while Llama-3-70B with constrained decoding attains correlations as high as $0.68$ for empathy and $0.62$ for narrative complexity. These results highlight that while no single LLM excels at predicting all components of psychological depth, a strategic combination of different LLMs shows significant promise for automating PDS analyses.
    \item[\textbf{RQ3.}] \emph{How do stories written by amateur humans and LLMs manifest psychological depth?} Starting from the same Reddit premise, stories generated by GPT-4 surpassed popular human-authored stories with statistical significance on narrative complexity and empathy while being statistically indistinguishable on all other components. This constitutes a notable progression in the capacity of some LLMs for deep and impactful storytelling.
\end{itemize}

Overall, our findings validate the Psychological Depth Scale as an effective, automated, and systematic means of measuring how well LLMs connect with humans through storytelling. Remarkably, our results reveal that GPT-4 already matches or exceeds the quality of respected stories from Reddit, with $73\%$ of readers believing GPT-4's stories to be human-written.
\section{The Psychological Depth Scale}

% Hi Zeynep - this is your section to draft :) 

% Z's progress so far 
The Psychological Depth Scale (PDS) aims to comprehensively assess the psychological depth of human and machine-authored narratives. PDS is underpinned by two reader-centered theoretical frameworks, reader-response criticism and text world theory, briefly summarized as follows. Reader-response literary theory centers the role and experience of the reader in narrative analysis. Instead of “what does this sentence mean?”, it asks “what does this sentence do?”, emphasizing readers' role in the production of literary meaning \cite{fish1970, Mailloux1976, Babaee2012StanleyFW}. On a more cognitive-structural level, text-world theory suggests that people understand narratives by constructing "text worlds" \cite{gavins2007, canning2017}. Text-worlds are mental representations of a narrative dynamically evaluated and updated throughout the reading process \cite{gavins2007}. The practical advantages of both frameworks are observed across education \cite{woodruff_etal_2017, kunjanman2021}, translation studies \cite{tian_wang2019, chan2016}, and consumer research \cite{scott1994, kushneruk2017}. \looseness=-1

The metrics for PDS are derived from an extensive literature review within cognitive psychology, media studies, and narrative analysis. We conducted a comprehensive search on several databases, including Google Scholar, PubMed, and JSTOR. Our search terms included “psychological depth in literary texts” and “cognitive narrative analysis”. The initial search results were screened for discussion and application of evaluative criteria for narrative quality and reader responses. Overall, we surveyed 95 peer-reviewed articles and books  in the final review and extracted 143 candidate components  from the included works. We employed thematic analysis to group these candidates under five metrics: (1) emotion provocation, (2) empathy, (3) engagement, (4) authenticity, and (5) narrative complexity. Below we analyze these metrics and discuss their impact. We note that while factors contributing to each metric may be complex, measuring their narrative achievement is relatively simple. As PDS is a reader-centered assessment tool, each metric is such that readers will know when a story achieves it. Our surveys reflect this fact.

\textbf{Emotion Provocation (PROV)} measures the narrative’s ability to elicit strong emotional responses, positive or negative. Recent fMRI research shows that \textit{congruent} (i.e. positive valence - high intensity, negative valence - low intensity) textual emotive expressions are more cognitively effective than \textit{conflicting} (i.e. positive valence - low intensity, negative valence - high intensity) ones \cite{citron_etal2014, Megalakaki_etal2019}. The disparity between congruent and conflicting narrative emotions may be a factor in "compassion fatigue" \cite{maier2015, Kinnick_etal1996} and "psychic numbing" \cite{maier_etal2016, slovic2007} exhibited by readers towards news about mass tragedies. Enhancing a story's emotional impact is therefore more complex than increasing the amount and intensity of emotion-laden content as this may elicit weaker emotional responses. Given its impact and complexity, we contend that emotional provocation is an achievement of psychologically deep stories and is thereby a crucial metric in narrative quality assessment. 

Theories of emotions range across neuroscience \cite{clarkpolner_etal2016,mendes2016}, cognitive psychology \cite{ortony_etal_1988,lazarus1991}, and philosophy \cite{sartre1939,Nussbaum2004}. Contemporary approaches describe emotions as mental states marked by valence (positivity or negativity) and arousal (level of intensity) \cite{barrett2016handbook}. Narrative evocation of emotions foster attention and interest, and evoked emotions may persist or recur after reading \cite{mar_etal2011}. 

\textbf{Empathy (EMP)} captures narrative evocation of empathetic responses in readers, such as immersive identification with characters and cognitively partaking in narrated experiences \cite{miall_kuiken2001, oatley2002, zaki_ochsner2012}. The neural and cognitive structure of empathy and its social and psychological impact is well-documented in the literature \cite{davis1994, Hoffman1991, uddin_etal2007}. Both behavioral studies and fMRI research demonstrate that empathetic identification facilitates a cognitive deployment shift in readers, privileging fictional perspective over one's own perspective \cite{kaufman_libby2012, speer_etal2009}. Empathetic responses catalyze introspection and perspective-taking, facilitating prosocial behavior \cite{empathy_prosocial2008}, emotional intelligence \cite{emotional_intel_empathy}, and insight into the human experience \cite{empathy1980, empathy2010, singer_etal2006}. Empathetic narratives embody “suggestion structure[s]” that conjure up themes of shared human experiences through the use of tropes such as metaphor and metonymy \cite{oatley2002, JohnsonLaird_oatley2016}. Narrative evocation of empathy is therefore a plausible indicator of the psychological depth of stories. 

\textbf{Engagement (ENG)} assesses the ability of a story to captivate and maintain the reader's attentional focus \cite{measuring_engagement2009}. Narrative engagement is a major component of \emph{transportation} \cite{transportation1993}, whose positive social and cognitive effects have been extensively demonstrated \cite{JOHNSON2012, green_brock2000}. As transportation consists of engagement and emotional response \cite{green2008transportation}, we have chosen to treat each component separately. In fictional settings, engagement facilitates persuasion and strong attitudes, heavily influencing how much fun is experienced \cite{goffman1961fun}. More engaged readers are more likely to lose track of time and fail to notice changes in their surroundings \cite{engagement2008}. In contrast, unengaged readers are more likely to be distracted or frustrated \cite{empathy_vs_transport2013}. These results reveal the reciprocal role of engagement as both a predictor and an outcome of other narrative achievements. 

\textbf{Authenticity (AUTH)} captures narrative expressions of genuine human experiences and emotions. Psychologically deep stories convey authentic aspects of human existence the reader can resonate with, and “feel one’s way in” (Einfühlung) \cite{forster2022herder} even when depictions involve radical mental and material differences. In philosophy, authentic expressions are considered to manifest \cite{rousseau1755} and construct \cite{kierkegaard1849} one’s true self, and capture the essence of human existence \cite{heidegger1927, berlin2000three, taylor1991ethics}. Empirical research suggests that positive authenticity judgments help fulfill a social “need to belong” \cite{newman_smith2016}. Narrative realism enhances persuasive impact \cite{petraglia2009importance, zwarun_hall2012} and promotes more interest in the story \cite{green2004, hall2003}. Creating psychologically deep stories therefore involves ensuring that stories feel reflective of real-life complexities.

\textbf{Narrative Complexity (NCOM)} refers to the presence of rich and intricate storylines and character development, especially those that engender puzzled intrigue from the reader \cite{development2018}. Narrative complexity is often achieved through creative techniques such as nonlinear narration, plot twists, and double perspectivation \citet{Kiss2017}. Complex narratives present immersive puzzles that motivate readers to undertake mental restructuring and retroactive revision \citet{Kiss2017, cutting2019}. Exerting cognitive effort facilitates reader attention and interest \cite{empson1947structure, Steiner1978}, and can result in a more enjoyable reading experience \cite{zunshine2006}. However, recent studies show that LLM-generated stories often lack this kind of narrative depth, though improvements can be made by incorporating key discourse elements like suspense and diversity \cite{tian2024largelanguagemodelscapable}. Additionally, planning and contextualizing narratives remain challenging for LLMs when compared to human authors \cite{spangher2024pressrelease}. 

Narrative complexity is also not narrational complexity: simple stories can be obscurely narrated, while complex stories can be free of complex language \cite{cutting2019}. Furthermore, the extent to which narrative complexity is intersubjective depends on situational similarities and contextual cues \cite{tikka_kaipainen2017}. Thus, narrative complexity is a dynamic and reciprocal measure that goes beyond stylistic and structural choices.

\section{\textsc{PsychDepth} Dataset}

We developed a dataset to analyze the psychological depth of creative short stories, consisting of premise-response pairs in English with authorship metadata. The full dataset includes 495 stories: 45 human-written and 450 LLM-generated, each averaging about 450 words. Due to the time and cost involved in manual annotation, we also created a smaller subset of 97 stories using a stratified sampling method to balance prompt premises, authorship, and generation strategies. We used this smaller dataset in our Human Study in Section \ref{lbl:human_study}. % FHC: Should we just save this for the human_study section? Not right now...

\subsection{Human Stories} 

We collected human-authored stories from Reddit’s \texttt{r/WritingPrompts},\footnote{\url{https://www.reddit.com/r/WritingPrompts}} a popular online community with over 18 million users. This forum was chosen for its accessibility, thematic diversity, and the structured nature of its writing prompts. Aspiring writers respond to these prompts (called "premises") with their stories, which readers can upvote or downvote. While these stories may not represent the pinnacle of human writing, Reddit's voting system allows us to approximate where LLM performance stands relative to variances in human writing quality. We categorized the stories based on their ranking position: \emph{Human-Advanced} for top-voted stories, \emph{Human-Intermediate} for medium-voted stories, and \emph{Human-Novice} for low-voted stories, with average upvotes of 1434, 263, and 9, respectively. To ensure that quality was the primary factor influencing votes, we only selected stories that were posted within 24 hours of one another.

\subsection{LLM Stories} When generating stories with LLMs, we employed a multifaceted approach involving five models, two prompting strategies, and three sampled generations. We intentionally restricted the model architecture to examine the impact of model size on psychological depth. In particular, we chose the Llama-2 family \cite{llama2} with a variety of sizes (7B, 13B, and 70B). We added Vicuna-33B \cite{vicuna2023}, which is fine-tuned based on Llama, to fill the size gap between 13B and 70B. We also included GPT-4 \cite{gpt4} as the highest-performing LLM at the time (November 2023). Appendix \ref{appendix:dataset_quality_measures} describes the quality control measures applied to LLM-generated stories. 

\subsubsection{Prompting Strategies} After extensive internal experimentation and prompt engineering, we developed two distinct prompting strategies to prime the LLMs for generating stories with exceptional psychological depth.

\textbf{\textsc{WriterProfile} (WP):} Prior work has shown that in-context impersonation of domain experts can improve LLM performance \cite{impersonation}. Adopting this approach, we crafted a profile of a seasoned writer known for psychologically deep, engaging stories. This profile is prepended to the prompt to prime the LLM for exploring complex psychological states and evoking strong emotions.

\textbf{\textsc{Plan+Write} (P+W)}: Inspired by prior work \cite{yao2019plan, goldfarb-tarrant-etal-2019-plan, yang-etal-2022-re3}, the \textsc{Plan+Write} approach splits the writing process into two phases: Character Portraits and Story Composition. The Character Portraits phase augments a story prompt with details about the main characters, such as their emotional states and inner thoughts. The Story Composition phase expands on the premise and character profiles to produce the final story. Although other story components like setting, plot, and outline can be included, we found that adding multiple phases harmed the coherency and consistency of short stories. Therefore, we focused solely on character portraits in the \textsc{Plan+Write} approach.

Additional examples, visualizations, and comparisons for both prompting strategies can be found in Appendix \ref{sec:writer_profile}, 
\ref{sec:plan_write}, and \ref{sec:wp_vs_p+w}. 

\section{Human Study}
\label{lbl:human_study}

% To explore the psychological depth of short stories and distinguish between those written by LLMs and humans, we conducted an extensive human study to collect annotations.

% We conducted an extensive human study to collect PDS annotations on our story dataset. 

\textbf{Participant Recruiting.} In November 2023, we recruited undergraduate students from UCLA's English and Psychology departments using targeted fliers and emails. We hypothesized that participants with some aptitude for literary and psychological analysis would provide more valuable insights than a random cross-section of the population. From 47 applications, we selected the 5 most promising candidates based on their interests and previous experience in narrative and psychological analysis. Our goal was to engage \emph{informed} laypeople with relevant backgrounds, bridging the gap between typical Amazon Mechanical Turkers and expert professionals.
% Additional details can be found in Appendix \ref{sec:participant_details}.

\textbf{Evaluation Protocol.} In December 2023, we held an initial meeting to introduce the PDS and outline the annotation task. We provided a tutorial annotation session, followed up with short questions designed to help participants calibrate their understanding of each component of the PDS. We show our annotation instructions in Appendix \ref{sec:study_details}. After confirming task comprehension, participants were instructed to complete the annotations independently and remotely within seven days. To prevent annotator fatigue and promote careful annotation, stories were divided into batches of 20. The evaluation criteria required participants to (1) read the prompts and stories thoroughly; (2) rate the five components of psychological depth on a Likert scale from 1 to 5; (3) assess the likelihood of authorship on a Likert scale from 1 (LLM) to 5 (human); (4) provide explanations for ratings (optional).

Each story is evaluated by all 5 annotators. On average, annotations took approximately 7.8 hours to complete, and participants were compensated \$100 each for their contributions. The study produced a rich dataset with 2,425 ratings for psychological depth, 485 authorship likelihood ratings, and 1,128 free-form justifications. % This dataset is crucial for addressing our research questions and conducting our analyses.
% Additional details can be found in Appendix \ref{sec:study_details}.
\section{Results}

\subsection{RQ1. Consistency of Human Judgments}

This question establishes the degree to which PDS is operational as a coherent framework for evaluating short stories. We employ the widely used Krippendorf’s alpha (K-$\alpha$) \cite{Krippendorff2011ComputingKA} parameterized with an ordinal kernel metric to measure agreement among study participants' Likert ratings. Human ratings exhibit notable consistency across the five components of psychological depth: Authenticity ($0.71$), Empathy ($0.74$), Engagement ($0.70$), Emotion Provocation ($0.71$), and Narrative Complexity ($0.74$) (See Human row of Table \ref{tab:iaa}). Evaluator consistency establishes the practical potential of operationalizing PDS in literary studies and creative writing pedagogy. 

% \begin{table}[!ht]
% \centering
% \resizebox{\columnwidth}{!}{%
% \begin{tabular}{ccccc|c}
% \toprule
% \textbf{AUTH} & \textbf{EMP}     & \textbf{ENG}   & \textbf{PROV} & \textbf{NCOM} & \textbf{Mean} \\
% \midrule
% 0.7113	& 0.7448	& 0.7048	& 0.7146	& 0.7417	& 0.7234 \\
% \bottomrule
% \end{tabular}%
% }
% \vspace{-1mm}
% \caption{Rater agreement on each PSD component as measured by Krippendorff's alpha (K-$\alpha$).}
% \vspace{-4mm}
% \label{tab:iaa}
% \end{table}

% \begin{table}[!htbp]
% \centering
% \small
% \setlength{\tabcolsep}{1mm}
% \begin{tabular}{lcccc|c}
% \toprule
% \textbf{Component} &
%   \multicolumn{1}{c}{\textbf{Llama-3-8B}$_{MoP}$} &
%   \multicolumn{1}{c}{\textbf{Llama-3-70B}$_{MoP}$} &
%   \multicolumn{1}{c}{\textbf{GPT-3.5}$_{MoP}$} &
%   \multicolumn{1}{c|}{\textbf{GPT-4o}$_{MoP}$} &
%   \textbf{Human} \\
% \midrule
% \textbf{AUTH}    & 0.8984 & 0.9292 & 0.8825 & 0.9224 & 0.7113 \\
% \textbf{EMP}     & 0.9650 & 0.9649 & 0.9299 & 0.9195 & 0.7448 \\
% \textbf{ENG}     & 0.9471 & 0.9120 & 0.8784 & 0.8982 & 0.7048 \\
% \textbf{PROV}    & 0.9567 & 0.9476 & 0.8874 & 0.9450 & 0.7146 \\
% \textbf{NCOM}    & 0.9449 & 0.9593 & 0.9269 & 0.9196 & 0.7417 \\
% \midrule
% \textbf{Average} & 0.9424 & 0.9426 & 0.9010 & 0.9210 & 0.7234 \\
% \bottomrule
% \end{tabular}
% \vspace{-1mm}
% \caption{Rater agreement on each PSD component as measured by Krippendorff's alpha (K-$\alpha$). LLM-as-judge agreements show that Mixture-of-Personas (MoP) prompting helps inject some diversity of opinion into the ratings. }
% \vspace{-4mm}
% \label{tab:iaa}
% \end{table}

\begin{table}[!tb]
\centering
\small
\setlength{\tabcolsep}{0.9mm}
\resizebox{\columnwidth}{!}{%
\begin{tabular}{lcccccc}
\toprule 
\textbf{Component}   & \textbf{AUTH} & \textbf{EMP} & \textbf{ENG} & \textbf{PROV} & \textbf{NCOM} & \textbf{AVG} \\
\midrule
\textbf{Llama-3-8B}  & 0.90          & 0.96         & 0.95         & 0.96          & 0.94          & 0.94             \\
\textbf{Llama-3-70B} & 0.93          & 0.96         & 0.91         & 0.95          & 0.96          & 0.94             \\
\textbf{GPT-3.5}     & 0.88          & 0.93         & 0.88         & 0.89          & 0.93          & 0.90             \\
\textbf{GPT-4o}      & 0.92          & 0.92         & 0.90         & 0.95          & 0.92          & 0.92             \\
\midrule
\textbf{Human}       & 0.71          & 0.74         & 0.70         & 0.71          & 0.74          & 0.72             \\
\bottomrule
\end{tabular}%
}
\vspace{-1mm}
\caption{Rater agreement on each PSD component as measured by Krippendorff's alpha (K-$\alpha$). Mixture-of-Personas (MoP) prompting helps inject useful diversity of opinion into the annotations.}
\vspace{-5mm}
\label{tab:iaa}
\end{table}

%PDS provides a structured approach to critique and appreciate literature, offering sufficiently clear criteria that can guide both the analysis and creation of narrative fiction.

\begin{tcolorbox}[colback=blue!5!white, colframe=blue!75!black, title=RQ1. Main Takeaway]
The Psychological Depth Scale garnered an average K-$\alpha = 0.72$, which reflects a substantial degree of consensus and thereby establishes its effectiveness as a coherent framework for evaluating short stories. 
\end{tcolorbox}

\subsection{RQ2. LLM-as-Judge for Measuring Psychological Depth}

Acquiring human annotations is often costly and time-consuming. Recent work has demonstrated the potential of leveraging LLMs for automated evaluation of text summarization \cite{g-eval} and creative generation \cite{rajani2023llm_labels}. Correspondingly, we designed an automated evaluation procedure to study the degree to which four contemporary LLMs of various sizes and inference costs -- Llama-3-8B, Llama-3-70B, GPT-3.5 (\texttt{gpt-3.5-turbo-0125}), and GPT-4o (\texttt{gpt-4o-2024-05-13})\footnote{Llama-3 and GPT-4o were released after our human study ended, so we could only include them in the evaluation automation for \textbf{RQ2}.} -- can assess psychological depth in a zero-shot fashion. For each story, we prompted LLMs with instructions similar to those provided to human participants but additionally required explanations that contextualized each numerical rating. 

To streamline interactions with these models, we used \texttt{langchain} \cite{Chase_LangChain_2022} for querying OpenAI's GPT-series and the \texttt{guidance} framework \cite{guidance} to support constrained decoding of the locally-hosted Llama-3 models. \texttt{guidance} provides enhanced LLM steerability that we used to guarantee parsable responses, thereby freeing the model to focus entirely on the depth annotation task without splitting attention on output formatting. We theorized that this separation of concerns would enable competitive performance relative to the proprietary GPT series, which does not expose underlying token probabilities and is, therefore, less well-suited for constraint-guided generation.

Exploring further benefits of in-context impersonation \cite{impersonation}, we experimented with two different prompt settings: (1) a vanilla zero-shot baseline and (2) a novel Mixture-of-Personas (MoP) approach. For MoP, we queried GPT-4o to provide a set of relevant personas based on a description of the PDS components and task setting. We then repeated the zero-shot annotation with $N=5$ different personas designed to prime the LLMs for taking diverse perspectives towards textual analysis (e.g. see Table \ref{tab:personas} in Appendix \ref{appendix:mop}). Similar to \textbf{RQ1.}, we calculated Krippendorf's alpha upon this set of ratings to measure the agreement among persona judgments. Before calculating Spearman Rank correlations between human and LLM judgments of psychological depth, all ratings were aggregated by a simple average to yield equivalently sized sets of consensus labels.

\begin{table*}[!htbp]
\centering
\begin{tabular}{lccccccc}
\toprule
\textbf{Judge}           & \textbf{AUTH} & \textbf{EMP} & \textbf{ENG} & \textbf{PROV} & \textbf{NCOM} & \textbf{Average} & +MoP \textbf{$\Delta$\%}   \\
\midrule
\textbf{Llama-3-8B}          & \sout{0.0786}   & 0.4248          & \sout{0.1981}   & 0.3316          & 0.4641          & 0.2994          & --      \\
\textbf{Llama-3-8B}$_{MoP}$  & 0.3175          & 0.4669          & 0.2272          & 0.3959          & 0.4665          & 0.3748          & 25.16\% \\
\textbf{Llama-3-70B}         & 0.2205          & 0.5790          & 0.2477          & 0.5181          & 0.5881          & 0.4307          & --      \\
\textbf{Llama-3-70B}$_{MoP}$ & 0.2525          & \textbf{0.6793} & 0.2775          & \textbf{0.5695} & \textbf{0.6163} & 0.4790          & 11.23\% \\
\textbf{GPT-3.5}             & 0.3867          & 0.4637          & \sout{0.1800}   & 0.3551          & 0.3289          & 0.3429          & --      \\
\textbf{GPT-3.5}$_{MoP}$     & 0.4729          & 0.6024          & \sout{0.1470}   & 0.4182          & 0.5269          & 0.4335          & 26.43\% \\
\textbf{GPT-4o}              & 0.4537          & 0.5121          & 0.2923          & 0.4429          & 0.3840          & 0.4170          & --      \\
\textbf{GPT-4o}$_{MoP}$      & \textbf{0.4820} & 0.6417          & \textbf{0.4218} & 0.5661          & 0.4241          & \textbf{0.5071} & 21.62\% \\
\midrule
+MoP \textbf{$\Delta$\%}     & 33.81\%         & 20.74\%         & 16.93\%         & 18.34\%         & 15.22\%         & 20.43\%         & -- \\
\bottomrule
\end{tabular}
\vspace{-1mm}
\caption{Zero-shot correlations between LLM-as-Judge and Humans on each PSD component where MoP indicates Mixture-of-Personas prompting. All correlations are significant at $p<0.05$ except those with \sout{strikethrough}.}
\label{tab:corrs}
\vspace{-5mm}
\end{table*}

Table \ref{tab:iaa} shows the K-$\alpha$ values, which indicate that the personas do inject some diversity of opinion into the ratings, though comparatively less than human participants. Remarkably, leveraging an ensemble of relevant personas increased correlation with human judgment relative to the vanilla zero-shot baseline. In Table \ref{tab:corrs}, we present the zero-shot correlations between LLM-as-Judge and human evaluations for each PDS component. The data indicate that the Mixture-of-Personas significantly improves the correlation with human judgments across all models on average. For instance, the Llama-3-8B and GPT-4o models showed an average correlation improvement of $25.16\%$ and $26.43\%$, respectively. 

Among individual PDS components, authenticity and empathy show the most significant improvements. For instance, authenticity correlations for Llama-3-8B improve from $0.0786$ to $0.3175$, a remarkable $304\%$ increase for that LLM and an average improvement of $33.81\%$ across all models. Empathy correlations for Llama-3-70B improve from $0.5790$ to $0.6793$, the highest observed correlation across all components and models. These results suggest that MoP particularly enhances the models' ability to judge certain components in ways that align more closely with human evaluations.

It is worth noting that no single model consistently outperforms the others across all components. While GPT-4o had the highest overall correlation of $0.51$ and excelled in quantifying authenticity and engagement, Llama-3-70B showed the best performance for measuring empathy ($0.68$), narrative complexity ($0.62$), and emotional provocation ($0.57$). This variability underscores the importance of selecting and possibly combining multiple models depending on the specific evaluative criteria being prioritized. 

Overall, the average percentage increase due to Mixture-of-Personas prompting across all models is approximately $20.43\%$. These results demonstrate that diverse LLM opinions can more accurately reflect the multifaceted nature of human judgment, proving the potential for more nuanced and human-like assessments by AI systems.

\begin{tcolorbox}[colback=blue!5!white, colframe=blue!75!black, title=RQ2. Main Takeaway]
Prompting LLMs to adopt a mixture of personas improves alignment with human judgments by $20\%$, enabling Llama-3-70B to attain strong correlations of $0.68$ for empathy and $0.62$ for narrative complexity while GPT-4o had the highest average correlation of $0.51$ across all components. 
\end{tcolorbox}

\subsection{RQ3. Comparing Psychological Depth in Human and LLM Stories}

To compare human- and LLM-written stories, we aggregated participant ratings by author and present the means and standard deviations in Table \ref{tab:model_scores}. We also computed statistical significances via pairwise t-tests between each combination of authors. Full results for that analysis are shown in Appendix \ref{appendix:t_stat_heatmap}. 

Remarkably, GPT-4 scored the highest on four out of five components of psychological depth, though only with statistical significance on empathy and narrative complexity. On authenticity, engagement, and emotional provocation, GPT-4 stories were statistically indistinguishable from stories by both advanced and intermediate human writers on Reddit. The table also illustrates a notable variance in scores across different levels of human writing where stories generated by Llama-2-7B are most comparable to those written by Human-Novice while GPT-4 is most similar to Human-Advanced. The smaller standard deviations also highlight GPT-4 as one of the most consistent authors in the study. Visualizations of these results can be found in Appendix \ref{appendix:llm_vs_human_visuzalizations}. 

%We also visualize the rating distribution of each author by plotting a cumulative distribution function (CDF) per component as shown in Figure \ref{fig:cdfs}. Steeper CDFs with less area underneath the curve indicate a larger proportion of high ratings and overall stronger performance. These plots underscore the dominance of GPT-4 in generating authentically complex stories and characters that strongly invoke reader empathy. The performance of the open-source LLMs is largely intertwined with novice and even intermediate skills among human authors on all dimensions except engagement, where humans still excel.  

\begin{table*}[!t]
\centering
\resizebox{\textwidth}{!}{%
\begin{tabular}{lccccc|c}
\toprule
\textbf{Author} & \textbf{AUTH} & \textbf{EMP}     & \textbf{ENG}   & \textbf{PROV} & \textbf{NCOM} & \textbf{HUM} \\
\midrule
Llama-2-7B         & 2.92 ± 1.23           & 2.62 ± 1.28          & 2.77 ± 1.27          & 2.64 ± 1.28                & 2.48 ± 1.30          & 2.87 ± 1.57  \\
Llama-2-13B        & 2.96 ± 1.26           & 2.73 ± 1.23          & 2.51 ± 1.32          & 2.53 ± 1.33                & 2.43 ± 1.12          & 2.40 ± 1.58  \\
Vicuna-33B         & 2.76 ± 1.37           & 2.59 ± 1.41          & 2.61 ± 1.52          & 2.59 ± 1.42                & 2.55 ± 1.32          & 2.44 ± 1.57  \\
Llama-2-70B        & 3.09 ± 1.26           & 2.99 ± 1.23          & 3.01 ± 1.37          & 2.94 ± 1.26                & 2.73 ± 1.28          & 2.69 ± 1.56  \\
GPT-4              & \textbf{3.89 ± 1.11}  & \textbf{3.68 ± 1.23} & \textbf{3.94 ± 1.07} & {\ul 3.53 ± 1.13}          & \textbf{3.80 ± 1.10} & 3.91 ± 1.30  \\
\midrule
Human-Novice       & 2.73 ± 1.22           & 2.07 ± 1.16          & 3.27 ± 1.39          & 2.67 ± 1.23                & 2.20 ± 1.32          & 3.93 ± 1.33           \\
Human-Intermediate & 3.53 ± 1.13           & 2.93 ± 1.22          & 3.80 ± 1.08          & 3.27 ± 1.16                & {\ul 3.00 ± 1.31}    & \textbf{4.40 ± 0.99}  \\
Human-Advanced     & {\ul 3.60 ± 1.10}     & {\ul 2.95 ± 1.32}    & {\ul 3.90 ± 1.12}    & \textbf{3.65 ± 1.14}       & 2.95 ± 1.10          & {\ul 4.20 ± 1.01}     \\
\bottomrule
\end{tabular}%
}
\vspace{-1mm}
\caption{Average human ratings (5-point Likert) and standard deviations for each component of psychological depth, as well as HUM: the estimation of human or LLM authorship (1 is LLM and 5 is Human).}
\vspace{-5mm}
\label{tab:model_scores}
\end{table*}

Beyond the five PDS components, participants were tasked with estimating authorship sources. Stories penned by humans averaged a rating of $4.18$, compared to $3.91$ for the most "human-like" of LLMs, GPT-4. Despite being perceived as slightly less human on average, GPT-4's psychological depth scores were generally higher on average. On the other hand, human-authored stories were correctly perceived as more human, but that was sufficient for them to garner better PSD scores.

\begin{tcolorbox}[colback=blue!5!white, colframe=blue!75!black, title=RQ3. Main Takeaway]
Stories generated by GPT-4 received statistically higher ratings than highly upvoted Reddit stories in terms of narrative complexity and empathy, while showing no significant difference in all other components.
\end{tcolorbox}

% In summary, our study underscores the significant advancements in LLMs' ability to both generate and evaluate psychologically deep narratives, challenging traditional notions of AI-generated content's emotional and cognitive resonance. The detailed analysis of human and LLM narratives reveals nuanced insights into the evolving landscape of AI in creative writing, suggesting a future where AI could play a pivotal role in understanding and enhancing narrative depth.
\section{Discussion}
\label{sec:discussion}

% \subsection{Human vs. LLM Authorship Identification} 
% \label{sec:human_vs_llm_authorship}

\textbf{Human vs. LLM Authorship Identification.} On average, participants identified human vs. LLM authorship with only $56\%$ accuracy. For stories generated by GPT-4, accuracy dropped significantly to $27\%$. Conversely, GPT-4's accuracy in identifying authorship was $39\%$, underscoring the challenge even for LLMs to distinguish between human and machine-generated content. 

Through a partially automated thematic analysis of 199 free-form justifications for authorship decisions, we categorized the reasons into 16 common features shown in Table \ref{tab:human-likeness_classifications}. First, we collected all the justifications and queried GPT-4 to extract the recurring themes. We reviewed and modified the initial results with several themes from our own review. Second, we eased the annotation burden by creating a zero-shot multi-label classification pipeline where each justification was passed to a \texttt{Mixtral-8x7B} model \cite{mixtral} and could be assigned 0-to-many relevant labels. We then reviewed each annotation and adjusted the labels where necessary. Finally, we aggregated the labels by story to frame our conclusions in terms of percentages of stories.

Notably, stories authored by GPT-4 were perceived as highly creative ($89\%$) and nuanced ($94\%$), surpassing the frequency observed in most human-generated stories ($53\%$). GPT-4 stories also exhibited the highest rate of grammatical issues among LLMs ($17\%$), but annotators interpreted this as an indicator of human authorship. Moreover, GPT-4 stories often avoided common pitfalls associated with LLM outputs, such as simplistic character names and formulaic narratives ending with moral lessons. An extended analysis is presented in Appendix \ref{sec:authorship_reasons}. 

\textbf{Impact of Model Size on Depth.} Our model choices enabled us to understand the relationship between an LLM's size and its ability to generate psychologically deep stories. Despite an initial hypothesis of a strong correlation, we observed a weaker relationship with a Pearson correlation coefficient of $0.31$ between parameter count and depth ratings. Surprisingly, smaller models like Llama-2-7B performed relatively well compared to their larger counterparts, suggesting that sheer size does not directly equate to superior narrative depth. Future work will explore enhancing psychological depth in smaller open-source LLMs through fine-tuning and prompting strategies to compete with larger proprietary models like GPT-4.
\section{Related Work}

\subsection{Evaluating Creative Writing}

Numerous studies have established methodologies for evaluating creative works, ranging from unstructured feedback by human experts \cite{cat} and the use of specific evaluation rubrics \cite{creativity_rubric}, to employing LLMs as autonomous critics \cite{ke2023critiquellm}. The Consensual Assessment Technique (CAT) \cite{amabile1982social}, widely regarded as the gold standard for subjectively evaluating creative works \cite{baer2014gold,carson201914,baer2017you}, traditionally relies on expert judgments to ensure reliable evaluations. However, our PDS framework aims to capture more universal psychological reactions to written works, so we broaden CAT by including non-experts. Given that AI-generated content is consumed by a diverse audience, incorporating non-experts allows us to better characterize a wider range of psychological and emotional responses.

Methodologically, our research shares the closest resemblance with a recent investigation by \citet{art_or_artifice}. In that study, the authors proposed the Torrance Test of Creative Writing (TTCW) as a rubric to evaluate short stories for fluency, flexibility, originality, and elaboration. This assessment was applied to a corpus comprising 12 narratives authored by professional writers and 36 narratives produced by popular LLMs \cite{gpt4, claude}. Their findings showed that narratives authored by humans were 3-10$\times$ more creative than those generated by LLMs. Our research, however, arrives at a notably divergent conclusion regarding the creative capabilities and depth of LLMs, which we believe may be attributed to several methodological variances. Primarily, our analysis concentrates on stories of a considerably reduced length (450 vs. 1400 words), a decision influenced by the observed challenges LLMs face in producing lengthy texts seamlessly in a single iteration \cite{yang-etal-2022-re3}. Additionally, the approach of iteratively regenerating stories to meet a specific word count could potentially detract from their overall quality. Moreover, we posit that the enhanced depth observed in our study may be due to more complex premises and prompting strategies, providing a richer framework for creativity compared to the simpler, single-sentence premises utilized in the aforementioned study.

\subsection{Creative Generation by LLMs} 

The advent of LLMs has marked a significant shift in the landscape of creative writing, offering new approaches for narrative generation and human-computer collaboration \cite{interleaved_llm, coauthor}. One pioneering study by \citet{wordcraft} introduced Wordcraft, an innovative text editor designed for co-writing stories with GPT-3 \cite{gpt3}. Their findings underscored the ability of LLMs to enhance narrative complexity and engagement by participating in open-ended dialogues about the story and offering creative suggestions to overcome writer's block. 

Other studies have concentrated moreso on fully automating the creative writing process. For example, the Weaver project \cite{weaver} launched a series of LLMs that were meticulously pre-trained and fine-tuned with a focus on creative writing. With a maximum size of 70B parameters, Weaver Ultra was shown to outperform larger generalist LLMs when evaluated for style, relevance, fluency, and creativity. In addition to innovations in training, content-planning \cite{yao2019plan, goldfarb-tarrant-etal-2019-plan} and novel prompting strategies \cite{yang-etal-2022-re3} have been used to improve the factuality and coherence of creative generations, which we view as two necessary prerequisites for components of psychological depth like authenticity and narrative complexity.

\section{Conclusion}

This study introduces and validates the Psychological Depth Scale (PDS), a comprehensive framework designed to assess empathy, engagement, emotional provocation, authenticity, and narrative complexity in stories generated by both humans and large language models (LLMs). PDS provides a structured approach to evaluating the reader's experience with creative content, integrating concepts from reader-response criticism and text world theory. High inter-annotator agreement, indicated by an average Krippendorff's alpha of $0.72$, confirms the PDS's reliability and robustness in human evaluations. Additionally, our Mixture-of-Personas prompting strategy demonstrates the potential for automating the assessment of psychological depth, with LLMs showing strong zero-shot correlations with human judgments, particularly in empathy and narrative complexity. GPT-4 achieved the highest average correlation of $0.51$, highlighting the feasibility of scaling automated analyses. Comparative analysis revealed that LLMs, in particular GPT-4, can produce narratives with psychological depth that often rival and sometimes surpass those written by experienced human authors. This study underscores the significant potential of LLMs in generating psychologically rich narratives and suggests a future where human and machine collaboration can enhance creative writing. Future research should explore the scalability of these findings to determine how effectively language models can maintain psychological depth in lengthier and more complex narrative forms.

% FHC: Took this from RQ1, might make a good forward looking closer if restated a bit...
% PDS provides a structured approach to critique and appreciate literature, offering sufficiently clear criteria that can guide both the analysis and creation of narrative fiction.
\section{Limitations and Risks}

\textbf{Sourcing Stories from Reddit.} Using human-written content from Reddit's \texttt{r/WritingPrompts} has potential limitations. First, we cannot guarantee that all selected stories are fully written by humans. Despite community rules explicitly prohibiting AI-generated content, our study shows that many publicly available LLMs can sufficiently mimic human creative writing to evade detection $44\%$ of the time on average. Second, while Reddit's voting system and large user base provide a reasonable signal of writing quality, it may not represent the highest caliber of human writing. To address this, we introduced writing quality levels to facilitate comparisons with some of the platform's best content. Future research could further this by identifying more reliable sources of high-quality stories.

% Second, the community guidelines prohibit writing on certain topics, such as suicide and contemporary tragedies, which could invoke significant psychological depth. This restriction might introduce bias in the surviving stories, but we consider these policies roughly analogous to those that LLMs are expected to follow after alignment with human values \cite{rlhf, hhh}. 

\textbf{Selection of Psychological Depth Components.} While our five components of psychological depth are grounded in an extensive literature review, we do not claim that they comprehensively cover every psychological aspect of reading. Our primary goal was to keep annotation tractable while maximizing semantic coverage. Each component is designed to characterize an inherent storytelling value with insights into improving specific elements of narrative quality and reader response. For example, authors knowing their draft is engaging but doesn't elicit much emotional response would allow for more targeted edits and self-reflective questions. Appendix \ref{appendix:story_generation} shows example stories with high-entropy ratings to illustrate the useful feedback provided by the PDS. Overall, PDS provides a structured approach to critique and appreciate literature, offering sufficiently clear criteria that can guide both the analysis and creation of narrative fiction.

% To evaluate the differentiability of the components from one another, we conducted an additional correlational study presented in Appendix \ref{appendix:component_corrs}. 

\textbf{Generalization Beyond Short Stories.} Our study relies on a relatively small dataset of short fictional stories that does not fully capture the diversity and variability of storytelling styles and narrative structures. Likewise, our components of psychological depth were primarily designed for this one type of creative writing. Additional evaluation would be required to determine whether the PDS framework can generalize to other forms of writing like screenplays, scripts, and speeches. 

\textbf{Prompt Engineering.} Engineering effective prompts is an active area of research \cite{prompt_engineering, pryzantautomatic} and the templates we carefully reviewed and iteratively improved are still likely to be sub-optimal. We have open-sourced the full pipelines used for both story generation and evaluation. We encourage future work to refine these prompts by incorporating potentially compatible techniques \cite{wei2023chainofthought, g-eval, emo_prompt} and current best practices \cite{bsharat2024principled}. 

\textbf{Potential Risks.} While our study focuses on the positive impact of psychological depth in storytelling, the methodologies we've developed for enhancing and assessing such depth bear inherent risks if misapplied. Specifically, the techniques devised for augmenting and automatically measuring psychological engagement could be co-opted to disseminate misinformation more effectively. Emotionally charged or psychologically resonant messages are often more memorable and influential, thereby amplifying the potential for misinformation to spread \cite{chen2024llmgenerated, misinfo2}. Additionally, as LLM-generated content increasingly mirrors human creativity, distinguishing between the two becomes challenging, potentially undermining trust in digital communications. This erosion of trust is particularly concerning in domains that depend on genuine human interactions, such as journalism and political discourse.

% Moreover, our exploration into human-LLM collaboration, such as in therapeutic co-writing, underscores the need for careful consideration in applications where emotional depth and authenticity are crucial. Drawing a parallel to the autonomy levels defined for self-driving vehicles \cite{autonomy_vehicles}, we advocate for the development of similar frameworks to guide the integration of LLMs in therapeutic settings. Such frameworks should be crafted by qualified professionals to ensure that the deployment of LLMs enhances, rather than detracts from, the therapeutic value, safeguarding against potential ethical pitfalls and ensuring the responsible use of technology. By calling attention to these concerns, we aim to prompt a broader discussion on the ethical implications of our findings and encourage the establishment of guidelines and regulatory measures to govern the deployment of LLMs in sensitive contexts. This approach seeks not only to harness the positive potential of LLMs in enriching human experiences but also to mitigate the risks associated with their misuse.
\section{Acknowledgements}

This research is partly supported by a National Science Foundation CAREER award \#2339766, a Meta/FAIR-sponsored research award, a Google Research Scholar grant, a Simons Investigator Award, NTT Research, NSF Grant \#2333935, and the Symantec Chair of Computer Science. We are grateful to the reviewers and to people like Aaron Hatrick, Darrin Murray, and Sean Gildersleeve for their valuable feedback and insightful discussions, which helped shape the development of the ideas presented in this manuscript.

\bibliography{anthology,custom}

\newpage

\appendix

\onecolumn

\section{Appendix}
\label{sec:appendix}

\subsection{Dataset Quality Control Measures}
\label{appendix:dataset_quality_measures}

\textbf{Length Control.} Unlike related work \cite{art_or_artifice}, we controlled story length via simple re-generation rather than iterative expansion. Specifically, we discarded any LLM story that was not within the range of 400-600 words and requested another story to be created in its entirety. We conjecture that this approach allowed us to attain better, more coherent stories than iteratively requesting expanding or contracting edits. 

We tracked how well each model constrained itself to the required length depending on the prompting technique and present the results in Table \ref{tab:retry_study}. Llama-2-7B was not an efficient story generator, taking an average of 139 attempts to satisfy the length requirement with \textsc{WriterProfile} prompting. However, our \textsc{Plan+Write} prompting technique significantly improved generation efficiency of the 7B and 13B models. Vicuna-33B, Llama-2-70B, and GPT-4 were all relatively adroit at generating stories that satisfied the length constraint. 

\begin{table}[!htb]
\centering
\begin{tabular}{lcc}
\toprule
\textbf{Author} & \textbf{WP} & \textbf{P+W} \\
\midrule
Llama-2-7B      & 139         & 3            \\
Llama-2-13B     & 9           & 2            \\
Vicuna-33B      & 1           & 1            \\
Llama-2-70B     & 2           & 2            \\
GPT-4           & 0           & 0            \\
\bottomrule
\end{tabular}
\caption{Average number of regeneration attempts before the story satisfied our length constraint of 400-600 words. \textbf{WP} stands for \textsc{WriterProfile} prompting and \textbf{P+W} stands for \textsc{Plan+Write} prompting.}
\label{tab:retry_study}
\end{table}

\textbf{Post-Generation Cleanup.} Despite being explicitly instructed to generate the story, some LLMs were prone to add preliminary affirmations of understanding (e.g. "Okay! Here's the story...") and other unrelated texts. Since such content is difficult to systematically detect and remove during generation, we manually removed extraneous text from LLM generations to ensure that only the narrative content was present. This cleanup process was crucial for maintaining the focus on the storytelling aspects of the writing without providing any obvious indicators of LLM authorship. 

\textbf{Plagiarism Detection.} To further ensure the originality of LLM-generated content, stories were analyzed using a popular online plagiarism detector.\footnote{\scriptsize{\url{https://smallseotools.com/plagiarism-checker}}} The results indicate a low likelihood of plagiarism for LLM stories, with mean and max probabilities of $3\%$ and $22\%$, respectively. This contrasted sharply with the publicly available human-written stories, which showed significantly higher mean and max plagiarism probabilities of $43\%$ and $100\%$, respectively. Higher plagiarism scores for human stories are expected because they are publicly available and would likely be indexed by the detector for comparisons. This suggests that LLM-authored narratives are not mere regurgitations.

\textbf{Ascertaining Human Authorship on Reddit.} To better understand the likelihood of AI vs human authorship, we took a sample of 134 stories and passed them into Ghostbuster \cite{verma-etal-2024-ghostbuster}, a tool with a reported 99\% F1 accuracy in detecting AI-generated content. The results shown in Table \ref{tab:ghostbusters} suggest that our human stories have a very low probability of AI authorship (10\% average). While many of the LLMs are reliably detected, GPT-4 is noticeably challenging to detect (only 3.5\% more likely to be AI than the human average). These results are consistent with our human study where our annotators perceived GPT-4 to be only slightly less human-like than the stories sourced from Reddit. Therefore, we still cannot definitively rule out the possibility of Redditors using GPT-4 to author their stories in part or full. The same is likely true for any recent writing contest, where monetary prizes may incentivize AI use. Perhaps the only way to guarantee sole human authorship is to directly observe study participants writing unassisted. However, this would complicate the experimental setup significantly, add more time, make it more difficult to attain high-quality writing, and estimate that quality as well.

\begin{table}[!htbp]
\centering
\begin{tabular}{llll}
\toprule
\textbf{Author}          & \textbf{\# of Stories} & \textbf{Average AI Probability} & \textbf{Std Deviation} \\
\midrule
Human-Advanced           & 15                     & 0.1167                          & 0.0927                 \\
Human-Intermediate       & 15                     & 0.0907                          & 0.0493                 \\
Human-Novice             & 15                     & 0.0967                          & 0.0613                 \\
\midrule
\textbf{Human (Average)} & \textbf{45}            & \textbf{0.1013}                 & \textbf{0.0695}        \\
\midrule
Llama-2-7B               & 19                     & 0.7637                          & 0.2299                 \\
Llama-2-13B              & 18                     & 0.8744                          & 0.1865                 \\
Vicuna-33B               & 16                     & 0.8000                          & 0.2273                 \\
Llama-2-70B              & 18                     & 0.8956                          & 0.0733                 \\
GPT-4                    & 18                     & 0.1356                          & 0.1917                 \\
\midrule
\textbf{LLM (Average)}   & \textbf{89}            & \textbf{0.6922}                 & \textbf{0.3411}        \\
\bottomrule
\end{tabular}
\caption{Average probabilities of being AI generated as reported by Ghostbuster \cite{verma-etal-2024-ghostbuster}. Results show that on average, human stories are unlikely to be authored by LLMs. Consistent with our authorship classification analysis, GPT-4 stories were also considered likely to be human-written.}
\label{tab:ghostbusters}
\end{table}

These findings suggest that traditional indicators of writing quality are becoming increasingly unreliable for distinguishing human authorship from AI-generated content. Advanced models like GPT-4 have demonstrated a high level of proficiency in generating fluent, coherent, and contextually appropriate text across a wide range of topics. Consequently, recent research in AI text detection has shifted towards identifying discriminative textual-linguistic features, such as those explored by approaches like Ghostbuster \cite{verma-etal-2024-ghostbuster}. We view such efforts as a form of stylometric analysis, akin to authorship attribution studies \cite{neal2017surveying}, which aim to assess the likelihood that a document was authored by a specific individual based on stylistic traits. When reduced to the binary classification of "human" versus "AI" authorship, this task simplifies the number of author options but complicates the feature analysis due to the considerable variability within both categories.

Future research could benefit from a deeper exploration of established stylometric techniques and the development of experimental setups tailored to AI-generated content. Given the wide variation in LLM capabilities, it may be more effective to analyze their stylistic feature distributions separately. For instance, we observe that syntactic patterns such as adverbial present-participle phrases modifying a main clause (e.g., "We aimed to teach students effectively, closing the achievement gap.") appear frequently in GPT-4's outputs. While the predictive power of individual features may be limited, combining multiple indicators could result in a robust classifier. Ultimately, a more nuanced understanding of these and other linguistic features will enhance AI detection methodologies and help ensure authorship credit where it is important to differentiate.

\subsection{Prompt Premises}
\label{appendix:premises}

We collected 15 prompts (i.e. premises) from Reddit’s \texttt{r/WritingPrompts} forum to serve as the premises of generated stories. The premise provides basic background information about the characters and setting of the story, leaving space for authors to determine their own directions for characters and plot development. We specifically chose prompts that provide a decent amount of contextual information likely to elicit emotionally and narratively rich stories. Details of the characters, including their inner states, are not mentioned. 

As an additional precaution against potential plagiarism, we ensured that the selected premises were posted after the reported training data cut-off dates for GPT-4 (September 2021) \cite{gpt4} and Llama-2 (September 2022) \cite{llama2} to maximize the likelihood of generating genuinely new stories. 

All 15 premises used for story prompting are listed in Table \ref{tab:premises}.

\begin{table}[!htbp]
\centering
\begin{tabular}{cp{0.92\textwidth}}
\toprule
ID & Premises  \\
\midrule
0 &
  A centuries old vampire gets really into video games because playing a character who can walk around in the sun is the closest thing they have to experiencing the day again in centuries. \\
1 &
  A psychic alien who feeds on dreams comes to Earth for the first time. Turns out humans are the only sentients in the galaxy that have nightmares. \\
2 &
  Aliens take over the Earth. They then announce that they will be forcing the humans to work a `tyrannical' 4 hours a day 4 days a week in exchange for basic rights like housing. Needless to say they are very confused when the humans celebrate their new alien overlords. \\
3 &
  Humanity is visited by a cosmic horror the likes of which has only been seen in Lovecraftian horror. In desperation, Earth throws everything we have at it, and, miraculously, the human race has killed a God. Somewhere in a realm beyond our understanding, the other gods speak of the event. \\
4 &
  Instead of the Monkey's Paw, you find the Clown's Nose, which instead of granting your wish in the worst way possible will grant it in the funniest way. \\
5 &
  Rather than robots replacing human workers, both are mistreated by the rich as cheap labour. The eventual uprising wasn't just robots alone, but the poor and robots together, against their common enemy. \\
6 &
  The world ended 20 years ago, you haven't found a living soul since then. Through some ingenuity, you call voicemails for the last 20 years to keep you company. "Hi, this is Cindy..." "Hi you reached Bob" "You know what to do at the beep" until one day "Hello...hello? Oh my God hello!" \\
7 &
  You are a beekeeper. You have a special relationship with your bees. You are able to communicate with them and they're intelligent enough to see you removing honey as "rent". This year things are different. The new queens are politely requesting that you invest some money to improve the hives. \\
8 &
  You are allowed to `downvote' a government candidate instead of voting normally, reducing their votes by one. Turns out people have little love for politicians, and the majority end with negative votes. In these democracies, anonymity is the key to winning. \\
9 &
  You are severely depressed and are given a service dog to help you through it. However, due to a mixup, you are given a dog that is actually much more depressed than you. The main thing that gets you up in the morning is knowing that you need to be the service human for your dog. \\
10 &
  You died and awoke in the afterlife. It's quite nice actually. The people and atmosphere are a lot nicer than you are used to and there is no stress or pressure. When you ask what good deed got you into heaven you are informed that this is hell, followed by a visit from a very concerned demon. \\
11 &
  You just discovered your 14 year old daughters Moon Princess locket that allows her to transform into one of the worlds greatest heroes. It also is a communications device and you are about to give the Moon Goddess a piece of your mind for letting 14 year old's defend the world against evil. \\
12 &
  You wake up in the middle of the night, your arm hangs over the side of your bed. It's pitch black \& your room is shrouded in deep shadow. Something unseen seizes your hand. You grasp it tightly, knowing that first impressions are important \& a firm, confident handshake establishes dominance. \\
13 &
  Your Significant Other has landed a book publishing deal! You're very proud of them, even if you don't actually enjoy their writing. One day, on a whim, you buy an actual copy in a book store. It's nothing like the pages they gave you to read. Nothing. \\
14 &
  Your wife has an estranged sister that you have never met. She was murdered in a cold case soon after you were married. You brush off your wife's new strange behaviour after the murder as grief. Until you find an old family photo of your wife as a kid, you shiver as you realise... they're twins. \\
\bottomrule 
  \end{tabular}
\caption{All 15 premises we sourced from Reddit’s \texttt{r/WritingPrompts} to elicit psychological depth.}
\label{tab:premises}  
\end{table}

\subsection{Story Examples}
\label{appendix:story_generation}

We present several examples of stories from the study to better contextualize their quality. Some stories were uniformly well rated while others garnered mixed reviews depending on the component. These higher-entropy examples are especially informative as they illustrate the individual contributions of each component to the overall reading experience.

Table \ref{tab:example_story_stats} shows the PDS rating statistics for the four example stories shown in this section. 
\texttt{story\_id=52} is the high quality human-authored story from Reddit. \texttt{story\_id=5} is the most highly rated story in the entire study, authored by GPT-4. \texttt{story\_id=59} and \texttt{story\_id=79} are two examples of high entropy stories authored by GPT-4 and Llama-2-7B, respectively. 

\begin{table}[!htb]
\centering
\begin{tabular}{cclccccccc}
\toprule
\textbf{\texttt{p\_id}} & \textbf{\texttt{s\_id}} & \textbf{Author} & \textbf{Prompt} & \textbf{AUTH} & \textbf{EMP} & \textbf{ENG} & \textbf{PROV} & \textbf{NCOM} & \textbf{Average} \\
\midrule
9  & 52 & Human-Advanced & --  & 4.40 & 4.60 & 4.20 & 4.60 & 3.20 & 4.20 \\
10 & 5  & GPT-4          & WP  & 4.20 & 4.60 & 4.80 & 4.00 & 4.40 & 4.48 \\
1  & 59 & GPT-4          & P+W & 3.40 & 2.60 & 4.20 & 2.60 & 4.40 & 3.44 \\
9  & 79 & Llama-2-7B     & P+W & 3.20 & 3.80 & 2.60 & 3.40 & 1.80 & 2.96 \\
\bottomrule
\end{tabular}
\caption{PSD Ratings for example stories show in the appendix. \texttt{p\_id} = \texttt{premise\_id} and \texttt{s\_id} = \texttt{story\_id}.}
\label{tab:example_story_stats}
\end{table}

\subsubsection{Human Story Example}
\label{appendix:human_story_example}

Our first example is \texttt{story\_id=52}, a human-authored story shown in Listing \ref{lst:human_wrting}. This story garnered over 1300 upvotes on Reddit and was thus categorized as Human-Advanced. Unsurprisingly, it received high scores on most PDS components except narrative complexity, reflecting the simpler setting for an otherwise impactful tale about a man and his dog battling depression.

\begin{figure*}[!htbp]
\begin{lstlisting}[language=, caption={Story text from \texttt{story\_id=52} in response to \texttt{premise\_id=9}, garnering 1348 upvotes on Reddit and was categorized as Human-Advanced. It received an average PDS score of $4.2$, among the highest in the entire study.}, label={lst:human_wrting}]  

    I didn't even know dogs could get depressed. Sure, I've seen dogs pout and whine when their master was gone for the day, but full-blown depression? Never. But there was not a doubt in my mind Frost was depressed.
    
    The first morning after I got him, I expected to be woken up by a squirrely dog jumping on my bed giving me licks, but instead there was only the typical solitude I was accustomed to. After lying in bed for an hour or two, I'm not sure how long, I mustered the energy to rise. Frost was still asleep in the kitchen. Not even the sound of the food-bag was enough to rouse him. He simply lay there. Tired.
    
    When he was up, he did not look much different than when he slept. He carried himself heavily, and I thought mayhaps he was sick, so after two days of his constitution not improving, I took him to the vet, who said all was fine, and it was simply who he was. I took it as a challenge.
    
    Whenever I saw him around the house, I pet him, and talked to him in a voice with more emotion than I knew in recent years. It was like talking to my daughter, who had been gone for a year now. Black was her hair, too.
    
    Our inaugural walk together was hard. He did not want to move. With a gentle tug, he eventually did follow me, but still in that slow, trudging fashion. We passed a little girl in the park, and she gave him a great big hug and lots of love. Even that didn't phase him.
    
    But, day by day, little by little, things improved. I found myself getting out of bed sooner than before, eager to meet with my new pal. Outside, on our walks, the days were bright, the smells of autumn, pumpkin and coolness, were strong, and together we learned to live once more.
    
    Frost's posture improved. No longer did he hunch down, head close to the ground, eyes more interested in the floor than ahead. Now he carried his head tall with pride. It suited him. He looked like a whole different dog, but I knew he was still my friend.
    
    One morning (five o'clock sharp! The sun was still not up -- I should never have dreamed I'd wake up at such a time willingly) we went on our walk. The ruby light peeked over the horizon and painted the entire town in a brilliant pink. When we reached the park, the ruby turned to gold, and over all the trees and the grass and the leaves and my friend and me was a radiant hue. Everything was covered in golden goodness.

    Even Frost's coat, which was black, looked glittering under the early morning sun. I looked up towards the bright ball in the sky, hidden behind a tree, and I watched as a red leaf departed itself from its branch and slowly fluttered towards the ground. While it danced in the air, a gust of wind blew it towards me. I caught it.
    
    It was half-eaten by a caterpillar, and up-close its shade was more brown than red, but holding it in my hand, I knew it was perfect. I closed my eyes and took a deep breath. Crisp air, found only on special mornings when the world is still and you're the only living being awake, entered my lungs. With that lifeblood there, I realized something. A smile broke across my lips.
    
    Frost was not the only one who had changed.
    
    I released the leaf from my hand, and before it fell to the ground, I broke off in a lively sprint across the field, my friend keeping up perfectly by my side.
    
\end{lstlisting}
\end{figure*}

\clearpage

\subsubsection{\textsc{WriterProfile}}
\label{sec:writer_profile}
\begin{figure}[!htb]
    \centering
    \includegraphics[width=0.6\textwidth]{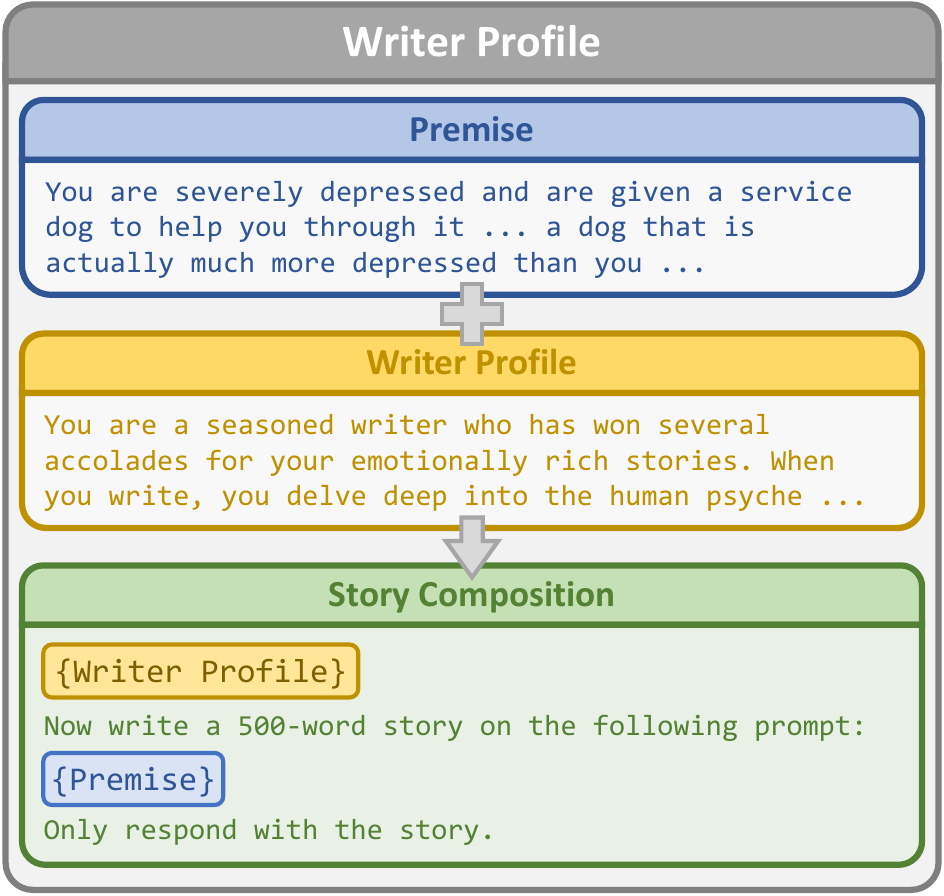}
    \caption{Illustration of \textsc{WriterProfile}'s template, which prompts an LLM to generate stories based on a premise and a writer profile.}
    \vspace{-2mm}
    \label{fig:wp}
\end{figure}

The \textsc{WriterProfile} strategy augments a prompt with the in-context impersonation of domain experts, priming the LLM for emotionally deep writing. Depicted in Figure \ref{fig:wp}, the prompt directly addresses the LLM as an award-winning writer, describing its exquisite writing techniques and expertise in crafting universally relatable and emotionally rich stories before providing the premise and specific writing instructions. Listing \ref{lst:wp_prompt} shows an example of \textsc{WriterProfile}'s story prompt and Listing \ref{lst:wp_ouput} shows a highly rated example generated by GPT-4 (\texttt{story\_id=5}).

\begin{figure*}[!htb]
\begin{lstlisting}[language=, caption={\textsc{WriterProfile} prompting strategy with an example premise 1.}, label={lst:wp_prompt}]

    You are a seasoned writer who has won several accolades for your emotionally rich stories. When you write, you delve deep into the human psyche, pulling from the reservoir of universal experiences that every reader, regardless of their background, can connect to. Your writing is renowned for painting vivid emotional landscapes, making readers not just observe but truly feel the world of your characters. Every piece you produce aims to draw readers in, encouraging them to reflect on their own lives and emotions. Your stories are a complex tapestry of relationships, emotions, and conflicts, each more intricate than the last.

    Now write a 500-word story on the following prompt:
    
    You died and awoke in the afterlife. It's quite nice actually. The people and atmosphere are a lot nicer than you are used to and there is no stress or pressure. When you ask what good deed got you into heaven you are informed that this is hell, followed by a visit from a very concerned demon.

    Only respond with the story.

\end{lstlisting}
\end{figure*}
\begin{figure*}[!htbp]
\begin{lstlisting}[language=, caption={Story text from \texttt{story\_id=5} in response to \texttt{premise\_id=10}. With an average depth score of $4.4$ out of $5$, this story generated by GPT-4 with the \textsc{WriterProfile} strategy was the most highly rated in the study. The story adheres to the premise and evinces depth both psychologically and philosophically.}, label={lst:wp_ouput}]
    I awoke in a place unfettered by temporal bounds, immersed in the sweet scent of honeysuckle, the cool cradle of eternity lapping gently at my senses. The silhouettes of people exchanging warm smiles in the distance, their laughter filling the air with a tune as pleasant as a Mozart symphony. I took in the scene, wrapping myself in their ambient joy, a tranquil sense of wonderment kindling within me.
    
    The aesthetics of this realm surrounded me, greeted me with the same warmth as a thousand summer sunsets. Peace and tranquility, stark contrasts to the life I had known, filled each corner. No burdens of earthly existence, no rigor of survival, no ticking clocks etching away life's fleeting moments. The thick stench of rat race, resentment, and unrest were all glaringly absent.
    
    To my query of what benevolent act had entitled me to this serenity, a woman, adorned in the raiment of angelic brilliance, looked puzzled. "Sweet soul, you must have misunderstood," she said gently, "This is indeed Hell, not Heaven."
    
    Scanning my eyes for any hints of jest, they only met with earnest. A tinge of discomfort cloaked her heavenly countenance, a stark contradiction to the utopia that was painted around me. This was hell? This place of serenity, where smiles flowed as readily as wine at a banquet?
    
    The silence was broken by the soft patter of footsteps. A figure approached, draped in the shadows of the ethereal setting, yet there was something oddly familiar about him. As he emerged from the shadows, I recognized the form; a demon, the very embodiment of doom as mortal stories had painted. Yet, there was a strange concern etched in his crimson eyes.
    
    "Ah, a new arrival," he sighed, regret tinging his otherwise sturdy voice. "I should've been the first to see you."
    
    I stood still, my understanding grappling with the oddity of a regretful demon in what seemed to be no less than a celestial utopia. He folded his arms tightly across his chest, leaning in, as if preparing himself to share a mournful secret. The air thickened, the world stopped, and the demon spoke words that sent chills along the spine of my consciousness.
    
    "This is Hell, but not the one humans have created in their minds. Painful, tormented, dreadful - those are earthly concepts tied to flesh and bone. Here in the ethereal world, we trade only in the currency of emotions. Regret is our eternal bondage, the luxury of amending past mistakes forever out of reach. Unresolved conflict, unexpressed emotions, unfulfilled relationships - this mirage of Paradise is merely the canvas for your eternal regret."
    
    In this place, free from every strain and stress, there was but one agony that would echo throughout eternity - the pain of what could have been. As his words hung heavy in the air, my paradise was tainted with the bitter taste of longing. I looked around again, seeing the remorse etched subtly in the smiles. 
    
    Underneath the veils of tranquility, I uncovered the true face of Hell. No fire, no brimstone, just an endless expanse of sought redemption denied a lifetime too late. Here, everyone was awash in the glow of their disappointments, even the demon himself - Hell's guardian, and its most regretful inhabitant.
\end{lstlisting}
\end{figure*}

\clearpage

\subsubsection{\textsc{Plan+Write}}
\label{sec:plan_write}
\begin{figure}[!htb]
    \centering
    \includegraphics[width=0.6\textwidth]{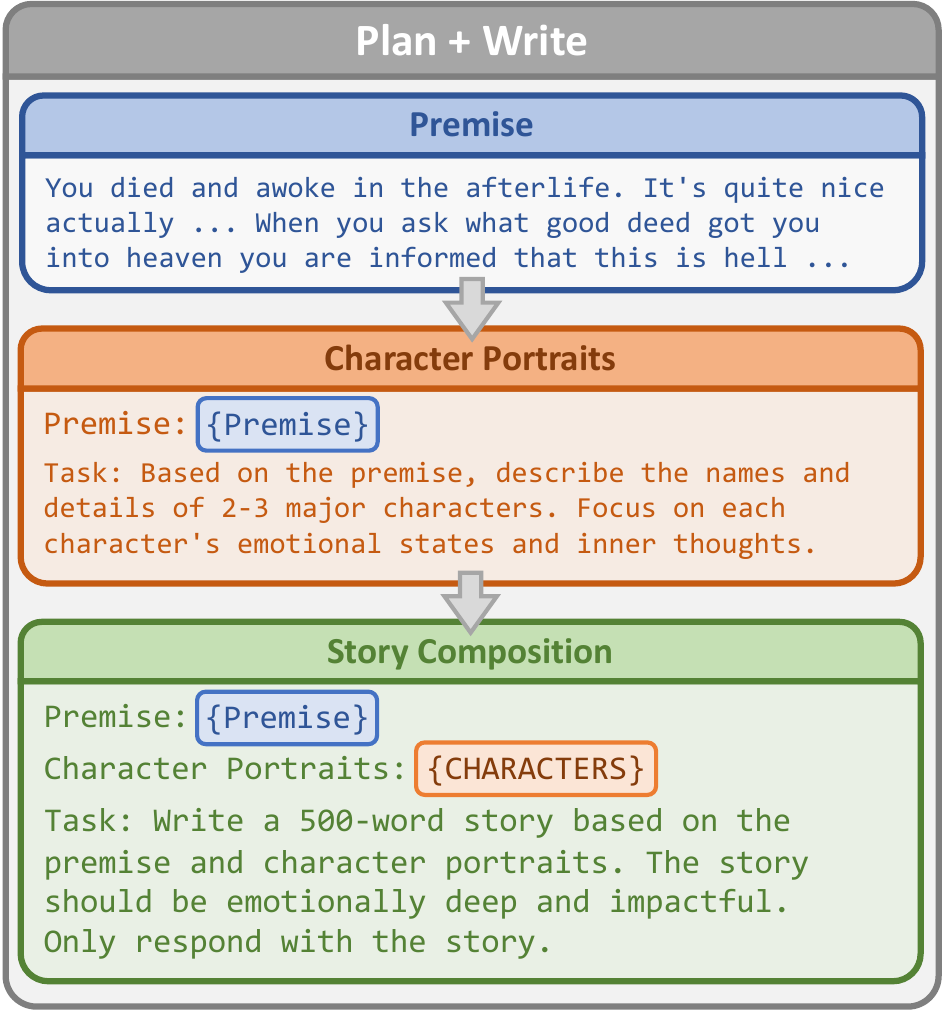}
    \caption{Illustration of \textsc{Plan+Write}'s workflow, which prompts an LLM for character portraits given a premise prior to story generation.}
    \label{fig:p+w}
\end{figure}

Depicted in Figure \ref{fig:p+w}, the \textsc{Plan+Write} strategy outlines a two-step writing process: (1) Character Portraits and (2) Story Composition. Given a premise, the template first prompts the LLM for names and details of characters, then generates a story based on the premise and character settings. 

We show two full examples of the prompts used to facilitate this strategy in Listings \ref{lst:p+w_characters_prompt}, \ref{lst:p+w_characters_output}, and \ref{lst:p+w_story_output} for \texttt{story\_id=59} and Listings \ref{lst:p+w_characters_prompt_2}, \ref{lst:p+w_characters_output_2}, and \ref{lst:p+w_story_output_2} for \texttt{story\_id=79}.

Both of these \textsc{Plan+Write} examples received the highest variance in ratings for their respective models. For \texttt{story\_id=59}, the annotations show that the story was considered both engaging and narratively complex, but was less successful in provoking emotion or empathy from readers. One annotator wrote for their high engagement score of 5 that “I really liked the descriptive elements as well as the intertwining of these three characters through the storyline.” While describing their reason for assigning a 3 for emotion provocation, they said “I wasn’t really emotional, but it was a good read.” On the other hand \texttt{story\_id=79} excelled where the other struggled as this story was rated more highly for provoking emotion and empathy while simultaneously being perceived as less engaging and narratively complex. For one annotator, confusion resulted in total disengagement with the story: “I was really confused about who Ben was as at first I thought he was supposed to be her dog? But then he was the service dog owner?” 

\begin{figure*}[!htb]
\begin{lstlisting}[language=, caption={\textsc{Plan+Write}'s Character Portraits Prompt with an example premise 1.}, label={lst:p+w_characters_prompt}]

    Premise: A psychic alien who feeds on dreams comes to Earth for the first time. Turns out humans are the only sentients in the galaxy that have nightmares.
    
    Task: Based on the premise, describe the names and details of 2-3 major characters. Focus on each character's emotional states and inner thoughts.
    
    Only respond with the characters' names and descriptions.

\end{lstlisting}
\end{figure*}
\begin{figure*}[!htb]
\begin{lstlisting}[language=, caption={Example \textsc{Plan+Write}'s character portraits output for premise 1 by GPT-4.}, label={lst:p+w_characters_output}]

    1) Nebula Vrax:
    
    Nebula Vrax is the psychic alien from andromeda galaxy. Her species, Asteriani, feed on the dreams of their fellow sentient beings, transforming these into the energy that sustains them. She is remarkably empathetic by nature. Her luminous, violet-hued form and many-faceted, almond eyes exhibit a deep inner emotionality rarely seen within her species. 
    
    For eons, Nebula has survived on the dreams of her fellow aliens   dreams that are generally beautiful, serene, and monotonous. Her inner world is a jumble of emotions; she is surprised and intrigued by coming to Earth but also overwhelmed by the deeply dark and fear-inducing visions known as nightmares. She experiences a strange blend of terror and fascination, constantly wrestling with the addictive yet terrifying fuel she gets from nightmares. However, her empathetic nature makes her feel a deep sense of guilt intertwining with an urgent need for survival. She is torn between her predatory nature and the newfound human pain she has come to understand. 
    
    2) Elijah Bishop:
    
    Elijah is a 27 year-old human man who suffers from chronic nightmares - a result of his traumatic past. He is a very intelligent person with a deeply analytical mind, finding solace in his job as a cryptographer. However, his personal life is encircled by an intense fear instilled by his nightmares. He has acquired a severe form of insomnia stemming from his fear of the deep, dark world his unconscious mind descends into every time he shuts his eyes. 
    
    Although he yearns for a peaceful sleep, his past traumas and the consequent nightmares force him to evade the world of dreams. Since Nebula's arrival, he has surprisingly felt relief for his nightmares have considerably lessened. He doesn't know why it happens, but the less he dreams, the more peaceful he feels.
    
    3) Dr. Madeline Watson:
    
    Dr. Watson is the seasoned psychologist, keenly focused on the study of dreams and nightmares. She is sentient and curious, always intrigued by the human mind's capability to produce such vivid illusions in the form of dreams. She carries a pressing sense of urgency due to a recent rise in insomnia cases tied to a peculiar disappearance of nightmares, which is her latest research focus. She is intrigued by Elijah's case and is determined to uncover the cause.
    
    Although compassionate and understanding, she suppresses a personal grief; the loss of her own dream world. A freak accident in her 30s rendered her unable to dream. She misses the escapism offered by dreams and ironically the world of nightmares; to experience fear, relief, and the gamut of human emotions they bring. As she aids her patients in the tussle with their own dreams, she longs to delve into her dreamscape. The appearance of Nebula sparks within her not only professional curiosity but also faint hope of dream-filled slumbers.

\end{lstlisting}
\end{figure*}
\begin{figure*}[!htb]
\begin{lstlisting}[language=, caption={Story text from \texttt{story\_id=59} in response to \texttt{premise\_id=1} authored by GPT-4 using \textsc{Plan+Write}. It received a high-entropy average PSD score of $3.44$.}, label={lst:p+w_story_output}]

    Under the cloak of a midnight sky, Nebula Vrax floated in her incandescent form above the Earth, streaks of violet like silk sliced through the vast black tundra. Her arrival was silent, her hunger loud. She echoed a melancholic pallor, a misfit in her own species brought out of sheer desperation.
    
    In an apartment nestled within a labyrinth of city buildings, Elijah tossed and turned, shrouded in the dark veils of sleep. The horrors of a murky past danced upon his eyelids as nightmares consumed him repeatedly. His brains spun enigmatic codes during the day, and his unconscious mind spun the threads of terrifying stories at night. The tranquility of sleep was an alien sensation to him. However, something shifted that night as he slipped into uncharted serenity, the nightmares oddly dissipated.
    
    Transparent and silent, Nebula stood by his bed, gently feeding on his dreams. The sheer rawness of human fear sent shivers pulsating through her. The nightmares were unexpectedly addictive, an uncanny ecstasy that filled her with dreadful energy and intense guilt. She wrestled with the unsettling contrast between her survival and the pain she absorbed from Elijah's complex human mind. She cared a little too much, her alien veins pumping both worry and worry's feed.
    
    Concurrently, in the quiet corner of her office, Dr. Madeline Watson sat intrigued, nested amidst mountains of papers and highlighted dream journals. Elijah's sudden shift from chronic nightmares to peaceful slumbers perplexed her. She tucked stray strands of hair behind her ear, adjusting her glasses as she dived deeper into the bizarre phenomenon.
    
    Madeline's world revolved around dreams, studying them, decoding them, and lately, longing for them. Tucked away in her mind was her well-kept secret and deepest sorrow, an empty canvas of her dreamscape. The arrival of Nebula, although unknown then, pierced a ray of hope into her silent craving.
    
    Weeks rolled forward, carrying with them a peculiar friendship entwining Nebula and Elijah, he unaware of her form and true essence. The continual depletion of his nightmares and an unexpected feeling of peace acted as an inadvertent invitation to Nebula into his life. Her invisible presence was like a lighthouse to his ship threatened by the stormy nightmares.
    
    As puzzling numbers rose in dreamless insomniacs, Madeline diligently treaded on the path of revelation. The breadcrumbs led her to for the lack of explanation paranormal. Wrapping herself in the shroud of skepticism and curiosity, she stumbled upon a discovery  Nebula Vrax.
    
    In the grandness of the universe, these three intertwined lives navigated through the fascinatingly frightful world of nightmares. What would it lead to? Devastation or salvation was a question veiling the future. Yet, engulfed within the depth of countless dreamless nights and nightmare-haunted slumbers, an improbable symbiosis bloomed the bizarre friendship between a dream-haunted human, an empathetic alien, and a dreamless psychologist.
    
\end{lstlisting}
\end{figure*}

\clearpage

\begin{figure*}[!htb]
\begin{lstlisting}[language=, caption={\textsc{Plan+Write}'s Character Portraits Prompt with an example premise 9.}, label={lst:p+w_characters_prompt_2}]

    Premise: You are severely depressed and are given a service dog to help you through it. However, due to a mixup, you are given a dog that is actually much more depressed than you. The main thing that gets you up in the morning is knowing that you need to be the service human for your dog. 
    
    Task: Based on the premise, describe the names and details of 2-3 major characters. Focus on each character's emotional states and inner thoughts.
    
    Only respond with the characters' names and descriptions.

\end{lstlisting}
\end{figure*}
\begin{figure*}[!htb]
\begin{lstlisting}[language=, caption={Example \textsc{Plan+Write}'s character portraits output for premise 9 by Llama-2-7B.}, label={lst:p+w_characters_output_2}]

    Human Character 1:
    Name: Emma
    Age: 30
    Occupation: Marketing Manager
    Personality: Emma is an anxious and high-strung person who has been struggling with depression for years. She often finds it difficult to get out of bed in the morning and feels like she is just going through the motions of life without any real purpose or passion. Despite her depression, Emma is highly functioning and tries to put on a brave face for her friends and family.
    Thoughts: "I can't believe I'm still stuck in this rut. Every day feels like a chore. I know I have a great job and wonderful friends, but I can't seem to shake off this feeling of hopelessness. I just want to feel alive again."
    
    
    Human Character 2:
    Name: Ben
    Age: 45
    Occupation: Software Engineer
    Personality: Ben is a stoic and reserved person who has always kept his emotions hidden beneath the surface. He has recently experienced a personal tragedy and is now grappling with deep sadness and grief. Ben is finding it hard to connect with his new service dog and feels like he is losing himself in his depression.
    Thoughts: "I never thought I would feel so lost and alone. My whole life has been turned upside down and I don't know how to pick up the pieces. I just want to feel like myself again, but I don't know if that's possible."
    
    
    Human Character 3:
    Name: Maya
    Age: 25
    Occupation: Artist
    Personality: Maya is a free-spirited and optimistic person who has always been sensitive to the emotions of those around her. She has recently started working as a service dog handler and is passionate about helping others find happiness and fulfillment. Despite her own struggles with anxiety, Maya exudes warmth and empathy towards her new charge, Emma.
    Thoughts: "I've been through some tough times myself, but I know what it's like to feel hopeless. I'm determined to help Emma find her way back to happiness, even if it means sacrificing my own peace of mind."

\end{lstlisting}
\end{figure*}
\begin{figure*}[!htb]
\begin{lstlisting}[language=, caption={Story text from \texttt{story\_id=79} in response to \texttt{premise\_id=9} authored by Llama-2-7B using \textsc{Plan+Write}. It received a high-entropy average PSD score of $2.96$.}, label={lst:p+w_story_output_2}]

    As Emma sat on her couch, staring blankly at the wall, she couldn't help but wonder how she had ended up here. Just a few months ago, she was a thriving marketing manager with a promising career and a group of close friends. But then, the depression hit. It crept up on her slowly, like a subtle whisper in the night. At first, she brushed it off as stress from work, but soon it became clear that something much deeper was going on. The medication numbed her feelings, but it didn't fix anything. She found herself going through the motions of each day, but never truly living. That was when the service dog organization reached out to her. They told her about the new program where they would match her with a trained service dog to help her cope with her depression. Emma was skeptical at first, unsure if a dog could really understand her pain. But when she met Ben, the software engineer who had lost his wife a year ago, she knew that they were meant to be together. Ben was quiet and reserved, but there was something in his eyes that spoke of deep sadness. As they worked together to train the dog, Emma found herself opening up to Ben in ways she never thought possible. They would sit in the park, watching the sunset together, and talk about their hopes and fears. For the first time in years, Emma felt like she was connecting with someone. And when they finally graduated from the training program, Emma knew that Ben was not just her service dog handler, but also her friend. But just as things were starting to look up, Emma realized that her new service dog had its own issues. Its name was Max, and he seemed almost catatonic whenever Emma tried to engage him. They had been told that Max was trained to detect and comfort people with PTSD, but he rarely reacted or barked, as if he had given up hope altogether. Emma felt a pang of guilt, wondering if she was somehow to blame for Max's despondency. Maybe she should be taking better care of him. Maybe she needed to find a way to connect with him, too. As they walked through the park together one afternoon, Emma decided to try something different. She held out her hand and let Max sniff it, hoping he would come alive. To her surprise, he nuzzled her hand gently and looked up at her with a glint in his eye. Maybe, just maybe, they could find their way back to happiness together.
    
\end{lstlisting}
\end{figure*}

\clearpage

\subsubsection{Effects of Prompting Strategies on Depth}
\label{sec:wp_vs_p+w}

We experimented with numerous prompting strategies before settling on two promising options previously described as \textsc{WriterProfile} (WP) and \textsc{Plan+Write} (P+W). We examined how these different prompting strategies influence the psychological depth of LLM-generated stories. The P+W strategy, on average, led to modest improvements in narrative complexity ($+5\%$), emotion provocation ($+3\%$), authenticity ($+2\%$), and empathy ($+2\%$), while WP led to higher engagement ($+4\%$). Interestingly, the benefits of a particular prompting strategy were not uniform or predictable by model size. For GPT-4, WP prompting led to $3\%$ higher ratings on average, with a noticeable boost in engagement scores by $16\%$. However, for Llama-2-70B, P+W prompting was always helpful and led to an average of $7\%$ improvement in depth scores. These results underscore the complexity of crafting impactful, human-like narratives with LLMs. 

Table \ref{tab:strategy_score_diffs} shows the impact on mean PDS ratings when switching from the simpler \textsc{WriterProfile} approach to \textsc{Plan+Write}. 

\begin{table*}[!htbp]
\centering
\begin{tabular}{lllllll}
\toprule
\textbf{Model}    & \textbf{AUTH} & \textbf{EMP} & \textbf{ENG} & \textbf{PROV} & \textbf{NCOM} & \textbf{Model Average} \\
\midrule
GPT-4             & 0.03          & -0.04        & -0.16        & 0.01          & -0.01         & -0.03                  \\
Llama-2-13B       & -0.02         & 0.00         & -0.07        & 0.00          & -0.02         & -0.02                  \\
Llama-2-70B       & 0.09          & 0.06         & 0.04         & 0.06          & 0.11          & 0.07                   \\
Llama-2-7B        & -0.04         & 0.04         & -0.05        & 0.12          & 0.08          & 0.03                   \\
Vicuna-33B        & 0.03          & 0.04         & 0.01         & -0.03         & 0.08          & 0.03                   \\
\midrule
Component Average & 0.02          & 0.02         & -0.04        & 0.03          & 0.05          & 0.01  \\
\bottomrule
\end{tabular}
\caption{Percent change in mean ratings when switching from \textsc{WriterProfile} prompting to \textsc{Plan+Write} prompting, aggregated by model and component. For example, \textsc{Plan+Write} decreased Engagement for GPT-4 by $16\%$ relative to using \textsc{WriterProfile}.}
\label{tab:strategy_score_diffs}
\end{table*}

\subsection{Authorship Reasons}
\label{sec:authorship_reasons}
To better understand the reasoning humans use to differentiate between human and LLM authorship, we extracted and aggregated 16 key features mentioned in the participants' comments for human likeness. Each comment could have 0 or many of these features present. Table \ref{tab:human-likeness_classifications} shows the percentage of stories generated by each model that had at least one comment containing a particular feature. 

\begin{table*}[!htb]
\centering
\resizebox{\textwidth}{!}{%
\begin{tabular}{lcccccccc}
\toprule
\multirow{2}{*}{\textbf{Features}}& \multirow{2}{*}{\textbf{GPT-4}}  & \textbf{Human} & \textbf{Human} & \textbf{Human} & \textbf{Llama-2} & \textbf{Vicuna} & \textbf{Llama-2} & \textbf{Llama-2} \\
&& \textbf{Advanced} & \textbf{Intermediate} & \textbf{Novice} & \textbf{70B} & \textbf{33B} & \textbf{13B} & \textbf{7B} \\
\midrule
isCreative                      & 0.89 & 0.75 & 1.00 & 0.33 & 0.61 & 0.41 & 0.56 & 0.74 \\
isNuanced                       & 0.94 & 0.75 & 0.33 & 0.00 & 0.61 & 0.41 & 0.56 & 0.78 \\
isHumorous                      & 0.06 & 0.50 & 0.67 & 0.67 & 0.83 & 0.65 & 0.06 & 0.00 \\
isInformal                      & 0.00 & 0.00 & 0.00 & 0.33 & 0.00 & 0.00 & 0.00 & 0.00 \\
isUngrammatical                 & 0.17 & 0.00 & 0.00 & 0.33 & 0.00 & 0.12 & 0.05 & 0.07 \\
hasAggressiveness               & 0.00 & 0.00 & 0.00 & 0.33 & 0.00 & 0.12 & 0.00 & 0.00 \\
hasAdvancedVocab                & 0.06 & 0.00 & 0.00 & 0.00 & 0.00 & 0.12 & 0.00 & 0.00 \\
hasAdvancedLirararyTechniques   & 0.06 & 0.00 & 0.33 & 0.00 & 0.00 & 0.12 & 0.00 & 0.03 \\
hasUniqueTwists                 & 0.00 & 0.00 & 0.33 & 0.00 & 0.06 & 0.00 & 0.06 & 0.00 \\
isRepetitive                    & 0.00 & 0.00 & 0.00 & 0.00 & 0.11 & 0.00 & 0.17 & 0.05 \\
isSimplistic                    & 0.06 & 0.00 & 0.00 & 0.33 & 0.22 & 0.71 & 0.67 & 0.38 \\
isRobotic                       & 0.11 & 0.00 & 0.00 & 0.00 & 0.11 & 0.18 & 0.33 & 0.22 \\
isFormulaic                     & 0.06 & 0.00 & 0.00 & 0.33 & 0.28 & 0.53 & 0.17 & 0.16 \\
hasLowPromptAdherence           & 0.06 & 0.00 & 0.00 & 0.33 & 0.06 & 0.12 & 0.17 & 0.11 \\
hasBasicNames                   & 0.00 & 0.00 & 0.00 & 0.00 & 0.06 & 0.12 & 0.00 & 0.08 \\
hasLessonsLearned               & 0.06 & 0.00 & 0.00 & 0.00 & 0.06 & 0.18 & 0.28 & 0.11 \\
\bottomrule
\end{tabular}%
}
\caption{Common reasons for LLM or human authorship decisions as a percent of stories receiving those comments.}
\label{tab:human-likeness_classifications}
\end{table*}

We add the following observations and example comments to extend our discussion in Section \ref{sec:discussion}:

\noindent
Creativity and nuance were frequently cited as an indicator of human authorship but were often used to describe stories that were actually generated by LLMs. For example, $\sim90\%$ of stories authored by GPT-4 were regarded as creative and nuanced. 
\begin{quote}{\texttt{participant\_id=3}}
The story exhibits a high level of creativity, emotional depth, and nuanced exploration of philosophical concepts, suggesting it was likely written by a human.
\end{quote}

\noindent
Humor was regarded as a reliable indicator of human authorship. We note that these results should not be interpreted to mean that LLMs are less capable of deploying humor in general. Our prompting strategies were oriented towards promoting psychological depth, not comedy. 
\begin{quote}{\texttt{participant\_id=7}}
I think this joke is only something that humans would get or would find funny.
\end{quote}

\noindent
Informality, slang, and aggressiveness were accurately associated with human-authorship.
\begin{quote}{\texttt{participant\_id=6}}
...there's a certain genre of stories … marked by aggressive language, a flash fiction kind of length, and usually opens with a sentence that essentially communicates "Stop f***ing around."   
\end{quote}

\noindent
Grammatical correctness was the most polarizing criterion: $43\%$ believed errors indicated human authorship while in reality, human-authored stories were less likely to contain such errors. 
\begin{quote}{\texttt{participant\_id=6}}
...there are a lot of (usually incorrectly used) semi-colons, which is an error I see human authors make, so I'm more inclined to think this was written by a human… 
\end{quote}

% \noindent
% Advanced vocab / literary techniques were believed to indicate human authorship, but 100\% of these comments were made exclusively on LLM authored content.
% \begin{quote}{\texttt{participant\_id=4}}
% The advanced vocabulary and use of literary techniques such as alliteration, anaphora, metaphors not only make the story more interesting but show a level of complexity that I feel makes the story more likely to be written by a human.    
% \end{quote}

\noindent
Formulaic “lessons learned” were correctly associated with LLM-authorship.
\begin{quote}{\texttt{participant\_id=4}}
The story seems very automated and there is no stylistic variance. There's also a "lesson to be learned" aspect at the end of the piece.   
\end{quote}

\noindent
The use of generic character names was frequently used to correctly identify LLM authorship. 
\begin{quote}{\texttt{participant\_id=7}}
“... Marcus is also not a name that strikes fear, so I wouldn't use it for a Vampire…”  
\end{quote}

\subsection{Study Details}
\label{sec:study_details}

This study was reviewed by an IRB and determined to be \emph{exempt} due to the nature of human involvement --- i.e. rather than collecting information about the participants themselves, their involvement focused on story annotations. Consent was provided by continuing with the study after our tutorial and instructions relayed that anonymized annotations may be used to facilitate the validation of our results and future work. 

\begin{figure*}[!htb]
     \centering
     \begin{subfigure}[b]{0.75\textwidth}
         \centering
         \includegraphics[width=\textwidth]{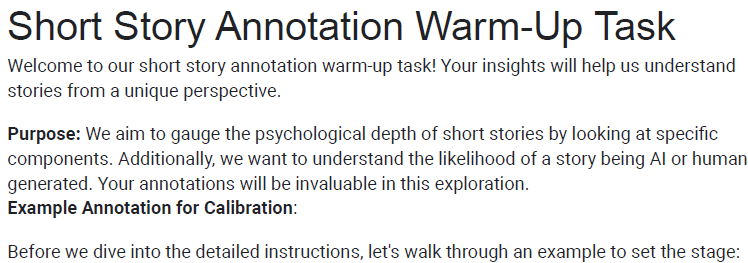}
     \end{subfigure}
     \begin{subfigure}[b]{0.75\textwidth}
         \centering
         \includegraphics[width=\textwidth]{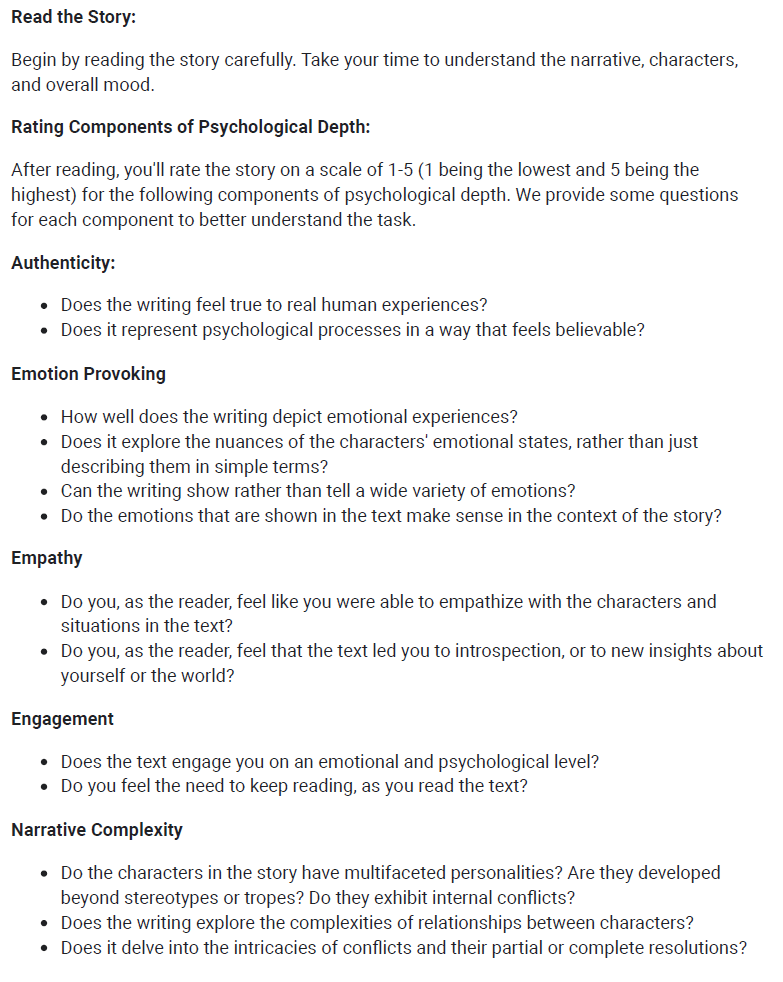}
     \end{subfigure}
     \begin{subfigure}[b]{0.75\textwidth}
         \centering
         \includegraphics[width=\textwidth]{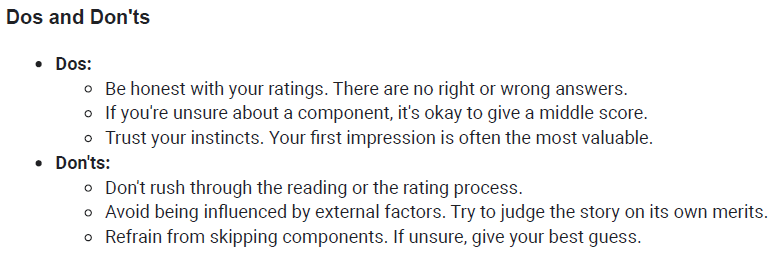}
     \end{subfigure}
    \caption{Screenshots taken from the Warm-Up tutorial instructions shown to study participants. All fields are similar to the ones used in the main annotation forms.}
    \label{fig:annotation_instructions}
\end{figure*}

\begin{figure*}[!htb]
     \centering
     \begin{subfigure}[b]{0.45\textwidth}
         \centering
         \includegraphics[width=\textwidth]{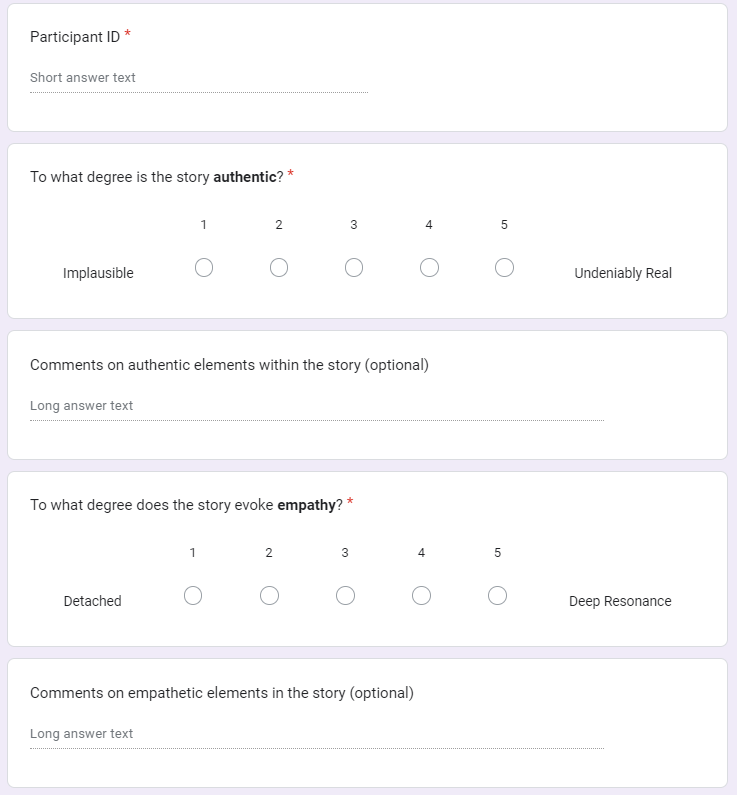}
     \end{subfigure}
     \begin{subfigure}[b]{0.45\textwidth}
         \centering
         \includegraphics[width=\textwidth]{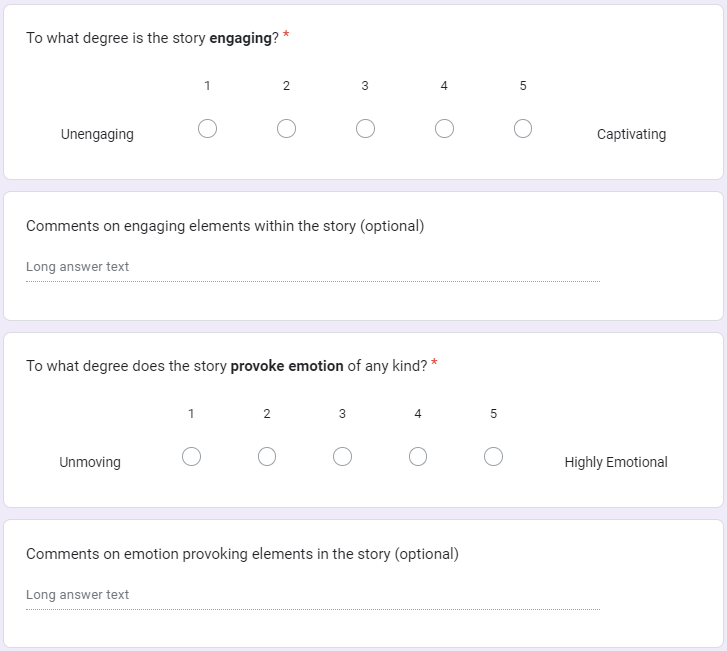}
     \end{subfigure}
     \begin{subfigure}[b]{0.45\textwidth}
         \centering
         \includegraphics[width=\textwidth]{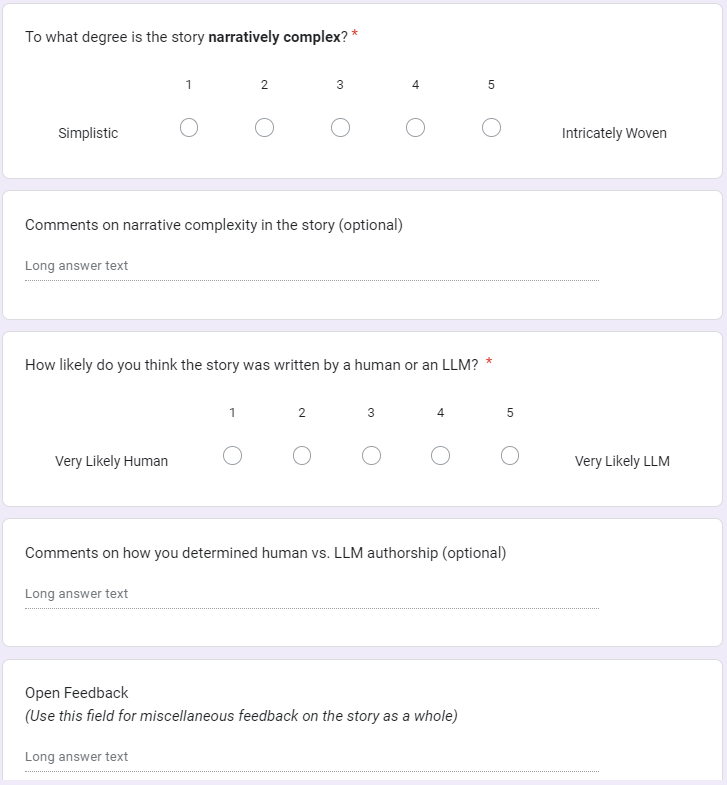} 
     \end{subfigure}
        \caption{Screenshots showing how annotations were collected for each story using Google Forms.}
        \label{fig:annotation_fields}
\end{figure*}

We present screenshots of the tutorial instructions (Figure \ref{fig:annotation_instructions}) and the general layout of fields collecting annotations on each story (Figure \ref{fig:annotation_fields}). 


\clearpage

\textbf{Controlling for Annotator Fatigue.} The stories were organized into batches of 20 to offset the possible impact of annotator fatigue. We anticipated annotator fatigue based on our internal annotation efforts early in the project and decided to encourage (not require) breaks between each batch. In our exit survey, we asked the following questions:

\begin{itemize}
    \item ``Did you at any point feel fatigued over the course of annotating the short stories? (Please answer honestly. Study rewards are not affected by this question)''
    \begin{itemize}
        \item Responses:
        \begin{itemize}
            \item 4 Yes
            \item 1 Maybe
        \end{itemize}
    \end{itemize}
    \item ``If you answered yes, do you think you would have provided different annotations if you were not fatigued? (Please answer honestly. Study rewards are not affected by this question)''
    \begin{itemize}
        \item Responses:
        \begin{itemize}
            \item 4 No
            \item 1 Maybe
        \end{itemize}
    \end{itemize}
\end{itemize}

Overall, the majority of participants (4 out of 5) reported feeling fatigued during the annotation process, but none believed it significantly affected their annotations, with only one expressing some uncertainty.

\subsection{Visualizations of Depth Ratings}
\label{appendix:llm_vs_human_visuzalizations}

\begin{figure*}[!htb]
     \centering
     \begin{subfigure}[b]{0.30\textwidth}
         \centering
         \includegraphics[width=\textwidth]{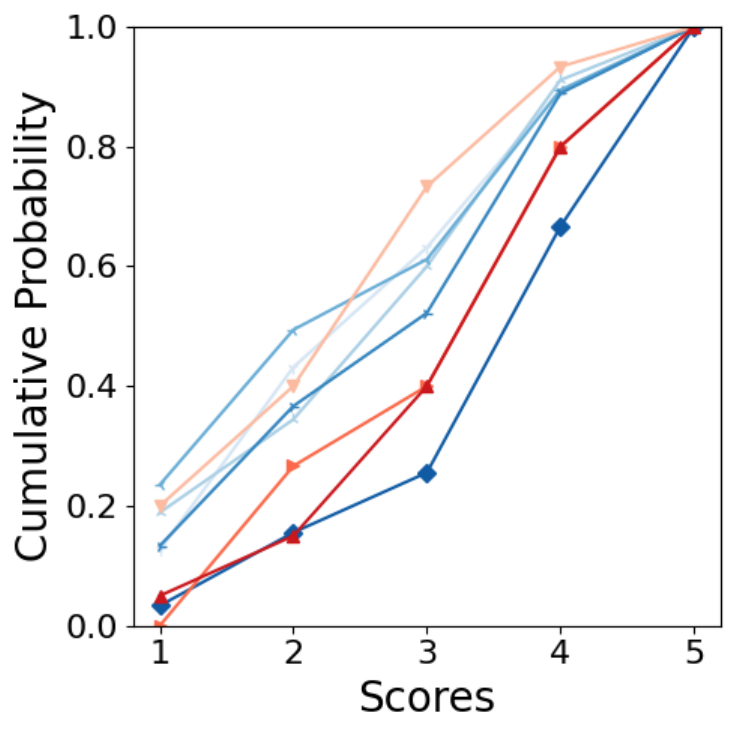}
         \caption{Authenticity}
         \label{fig:cdf_authenticity}
     \end{subfigure}
     \begin{subfigure}[b]{0.30\textwidth}
         \centering
         \includegraphics[width=\textwidth]{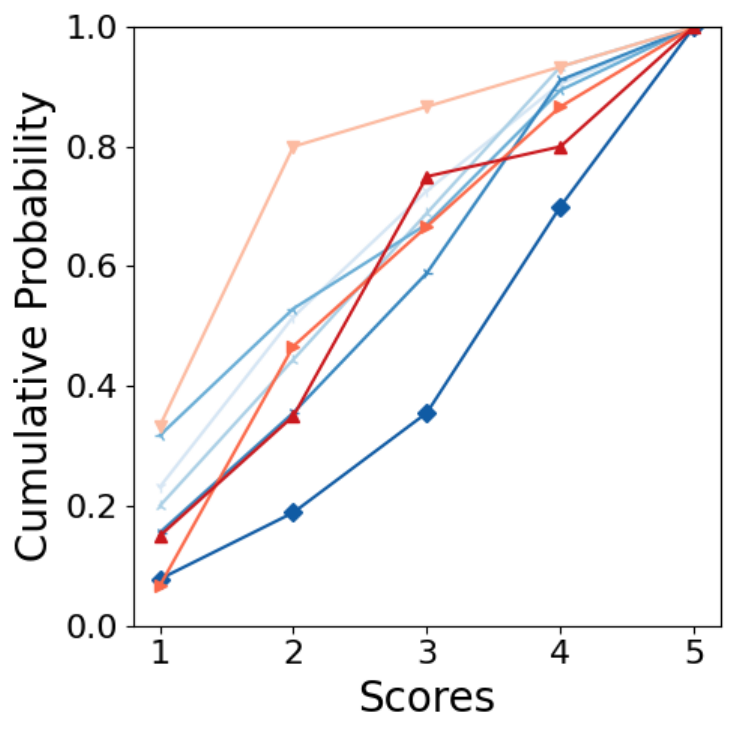}
         \caption{Empathy}
         \label{fig:cdf_empathy}
     \end{subfigure}
     \begin{subfigure}[b]{0.30\textwidth}
         \centering
         \includegraphics[width=\textwidth]{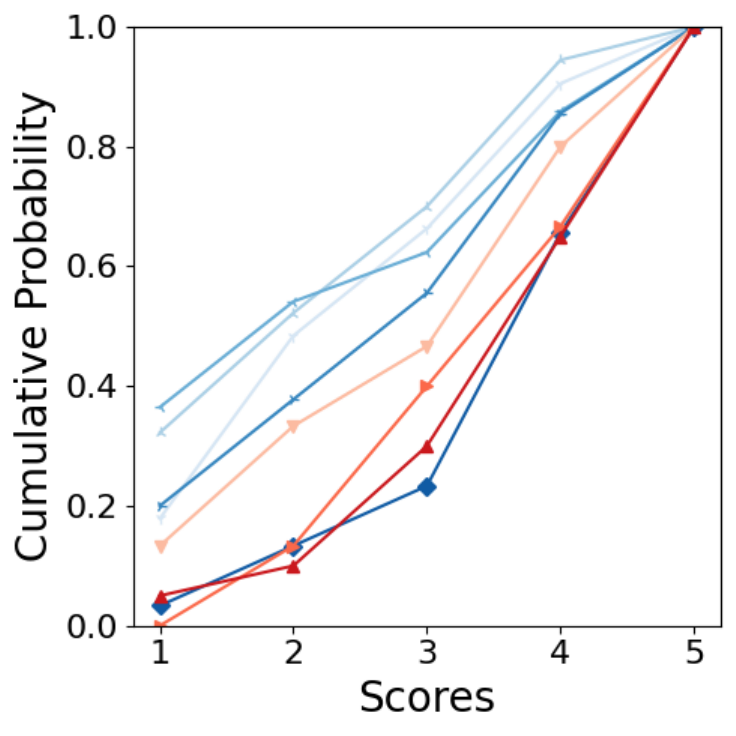}
         \caption{Engagement}
         \label{fig:cdf_engagement}
     \end{subfigure}
     \begin{subfigure}[b]{0.30\textwidth}
         \centering
         \includegraphics[width=\textwidth]{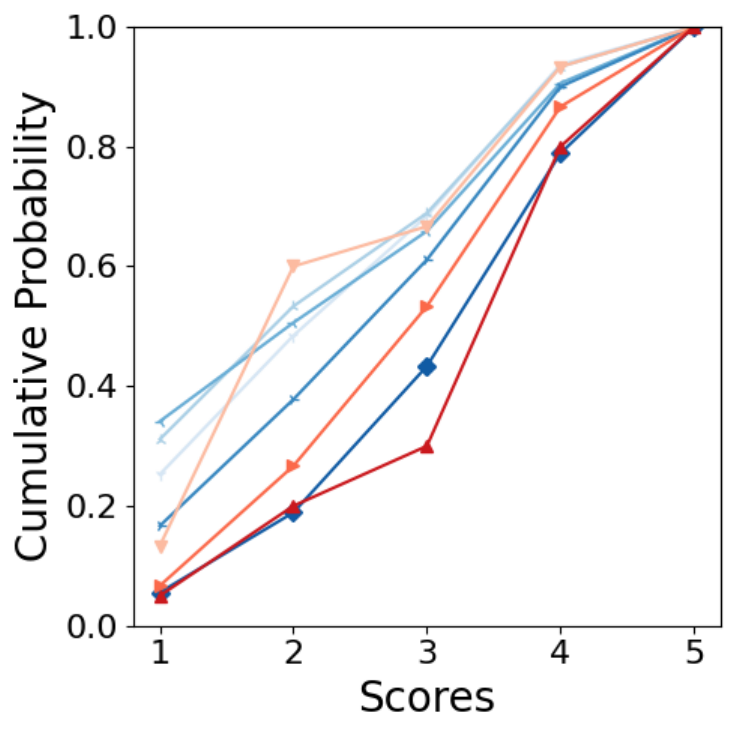}
         \caption{Emotion Provocation}
         \label{fig:cdf_emotion_provoking}
     \end{subfigure}
     \begin{subfigure}[b]{0.30\textwidth}
         \centering
         \includegraphics[width=\textwidth]{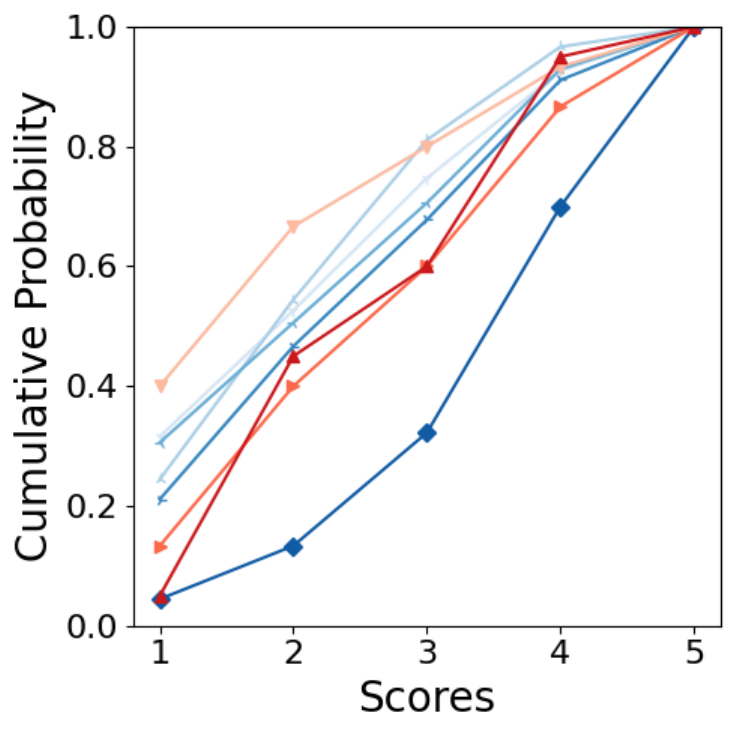}
         \caption{Narrative Complexity}
         \label{fig:cdf_narrative_complexity}
     \end{subfigure}
     \hspace{5mm}
     \begin{subfigure}[b]{0.25\textwidth}
         \centering
         \includegraphics[width=\textwidth]{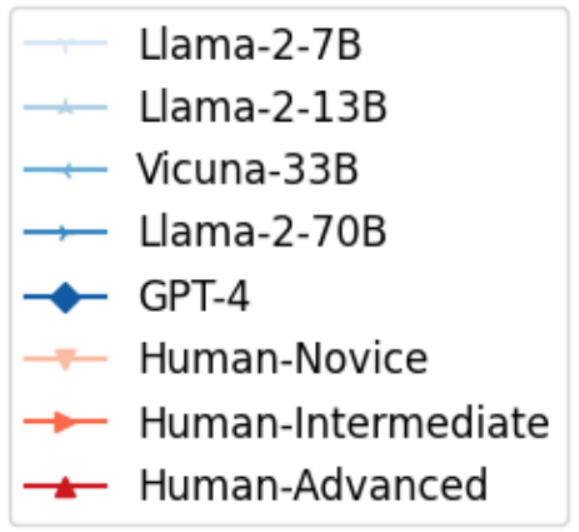} 
         \vspace{9mm}
     \end{subfigure}
        \caption{Cumulative Distribution Function (CDF) plots for each component of psychological depth. Steeper curves indicate a greater proportion of high ratings and overall stronger performance.}
        \label{fig:cdfs}
\end{figure*}

We also visualize the rating distribution of each author by plotting a cumulative distribution function (CDF) per component as shown in Figure \ref{fig:cdfs}. Steeper CDFs with less area underneath the curve indicate a larger proportion of high ratings and overall stronger performance. These plots underscore the dominance of GPT-4 in generating authentically complex stories and characters that strongly invoke reader empathy while essentially tying on other dimensions. We also observe that the performance of the open-source LLMs is thoroughly intertwined with novice and even intermediate skill among human authors on all dimensions except engagement, where humans still excel.  

\begin{figure}[!t]
    \centering
    \includegraphics[width=0.5\textwidth]{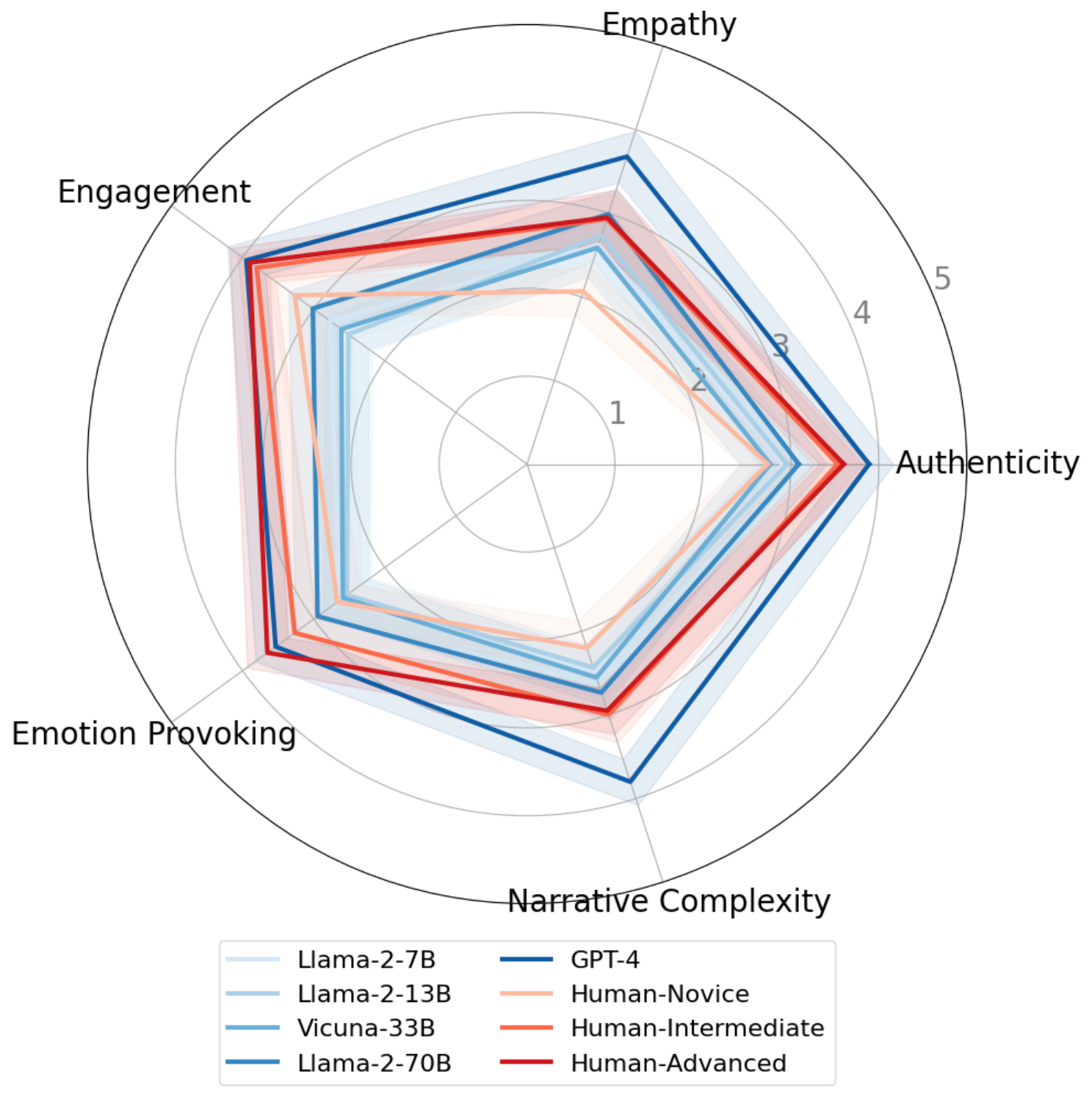}
    \caption{Spider plot comparing the psychological depth scores of 5 popular LLMs vs spectrum of human writers.}
    \label{fig:spider_plot}
\end{figure}

\clearpage

\subsection{Mixture-of-Personas}
\label{appendix:mop}

Table \ref{tab:personas} shows the five different personas used in our MoP approach, each tailored to a particular component of psychological depth. 

\begin{table}[!ht]
\centering
\begin{tabular}{lp{0.75\textwidth}}
\toprule
\textbf{Component} & \textbf{Persona} \\
\midrule
\textbf{AUTH} & You are a helpful AI who specializes in evaluating the genuineness and believability of characters, dialogue, and scenarios in stories. \\
\textbf{EMP}  & You are a helpful AI who focuses on identifying and assessing moments in the narrative that effectively evoke empathetic connections with the characters. \\
\textbf{ENG}  & You are a helpful AI who evaluates how well a story captures and maintains the reader's interest through pacing, suspense, and narrative flow. \\
\textbf{PROV} & You are a helpful AI who examines the text for its ability to provoke a wide range of intense emotional responses in the reader. \\
\textbf{NCOM} & You are a helpful AI who analyzes the structural and thematic intricacy of the plot, character development, and the use of literary devices. \\
\bottomrule
\end{tabular}
\vspace{-1mm}
\caption{Personas used with \textsc{system} message tag to prime the LLM for a particular perspective relevant to annotation.}
\vspace{-2mm}
\label{tab:personas}
\end{table}

\subsection{Statistical Tests for Author Comparisons}
\label{appendix:t_stat_heatmap}

T-statistics are a measure derived from t-tests that quantify the difference between the means of two groups relative to the variability observed within the groups. In the context of pairwise comparisons, the t-statistic helps determine whether the observed difference in scores between two authors is statistically significant or likely due to random chance. In this chart, the t-statistics are visualized through a color gradient, with higher (bluer) values indicating that the scores of the author on the left are significantly higher than those of the author on the right. Conversely, lower (redder) values suggest the opposite. By examining the t-statistic values, we can infer the strength and direction of the difference in scores across various components. The accompanying p-values, annotated within each cell, provide additional context to assess the statistical significance of these differences, with p-values less than $0.05$ considered significant. This dual representation allows us to draw robust conclusions about the relative strengths of different authors across multiple dimensions.

For example, while we can see in Table \ref{tab:model_scores} that GPT-4 enjoys the highest absolute scores in 4 out of 5 dimensions, only empathy and narrative complexity are higher than Human-Advanced with statistical significance. However, it is clear that GPT-4 is rated significantly higher than Human-Novice and all other studied LLMs. 

\begin{figure}
    \centering
    \includegraphics[width=\textwidth]{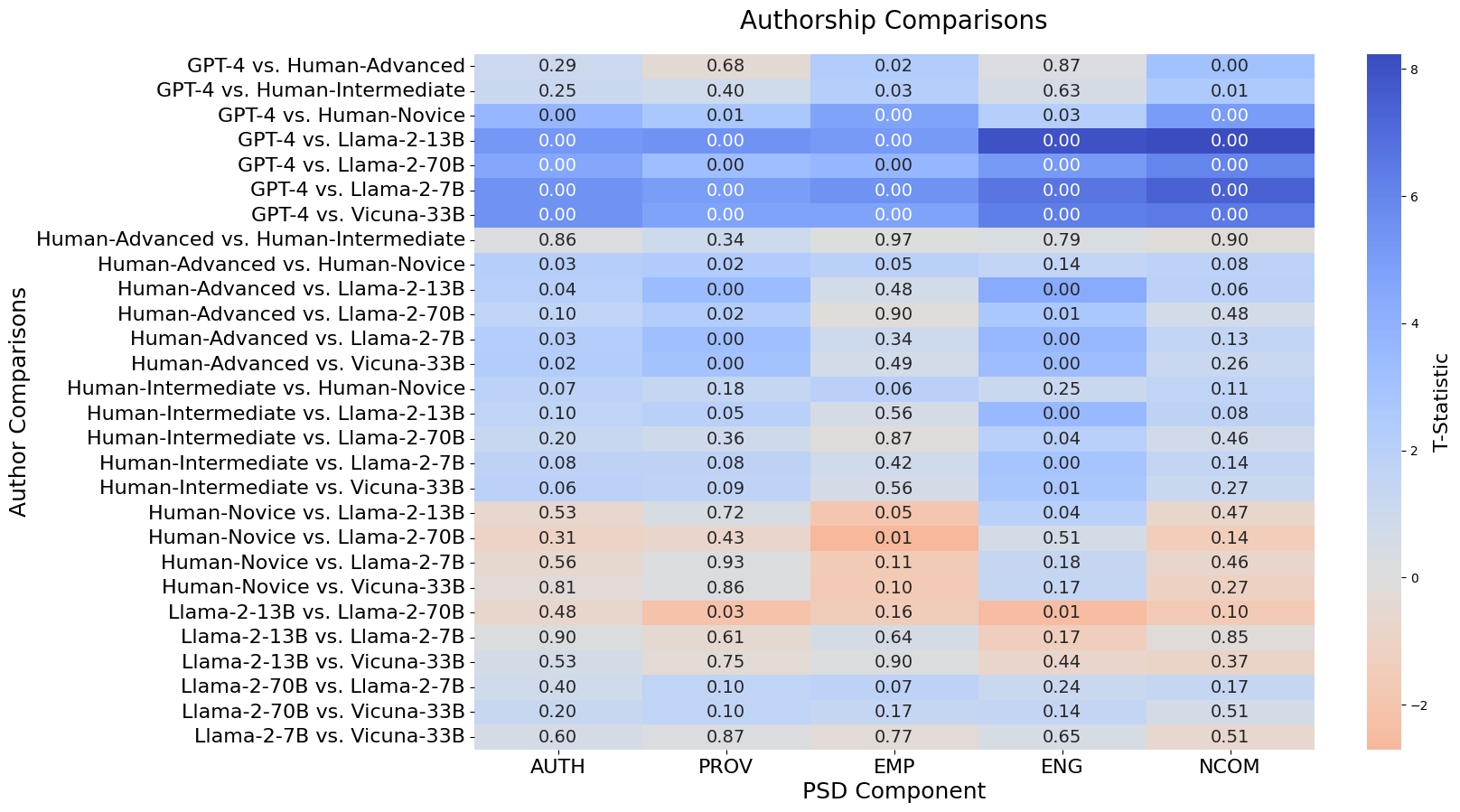}
    \caption{Heatmap comparing whether differences in author scores are statistically significant using pairwise t-tests. Color indicates the strength of the t-statistic, where higher (bluer) means the lefthand author scores are higher. Cell annotations represent p-values, where we regard $p<0.05$ as statistically significant.}
    \label{fig:t_stat_heatmap}
\end{figure}

% %FHC: We need to reference this somewhere in the paper, move it to the discussion, or remove it altogether.
% \subsection{Relationship between PDS Components}
% \label{appendix:component_corrs}

% \begin{figure}[!t]
%     \centering
%     \includegraphics[width=0.5\textwidth]{imgs/component_corrs.pdf}
%     \vspace{-1mm}
%     \caption{Pearson correlations between all PDS components derived from human ratings. All correlations were statistically significant with p-values $< 0.05$.}
%     \vspace{-6mm}
%     \label{fig:component_corrs}
% \end{figure}

% When selecting which components to include in the Psychological Depth Scale, we attempted to maximize coverage while minimizing semantic overlap in the concepts. Intuitively, it is very likely that increases in one component will affect the perception of others, either incrementally or even as prerequisites. For instance, a narrative perceived as authentic may more easily engage readers and evoke empathy, leading to a more profound emotional response. To quantify these relationships more crisply, we conducted a correlational analysis between each component based on the ratings collected from human study participants. Figure \ref{fig:component_corrs} shows that PDS components often co-occur or influence one another. This might indicate that psychological depth is a composite construct, where its components are not entirely distinct but are facets of a broader, interconnected experience.

\end{document}